\documentclass[journal]{IEEEtran}
\ifCLASSINFOpdf
\else
\fi

\usepackage{subfigure}
\usepackage{epsfig}
\usepackage{graphicx}
\usepackage{amsmath}
\usepackage{amssymb}
\usepackage{algorithm, algorithmic}
\graphicspath{{data/}}   % 设置图片所存放的目录``
\usepackage{multirow}

\usepackage{url}
\usepackage{hyperref}
\usepackage{booktabs}
\usepackage{color}
\usepackage{bbding}
\usepackage{tabularx}
\begin{document}
\title{ HyperDID: Hyperspectral Intrinsic Image Decomposition with Deep Feature Embedding}

\author{Zhiqiang~Gong,
			 Xian~Zhou,
       Wen~Yao,
       Xiaohu~Zheng,
       Ping~Zhong, ~\IEEEmembership{Senior Member,~IEEE}
        %Ping~Zhong,~\IEEEmembership{Senior Member,~IEEE}
        %and~Weidong~Hu% <-this % stops a space
\thanks{Manuscript received XX, 2023; revised XX, 2023. This work was supported by the Natural Science Foundation of China under Grant 62001502.}
\thanks{ Z. Gong, X. Zheng and W. Yao are with the Intelligent Game and Decision Laboratory, Defense Innovation Institute,
Chinese Academy of Military Sciences, Beijing 100071, China. e-mail: (gongzhiqiang13@nudt.edu.cn, zhengboy320@163.com wendy0782@126.com).}
\thanks{X. Zhou is with  the Information Research Center of Military Science, Chinese Academy of Military Sciences, Beijing 100000, China (e-mail: zhouxian@sjtu.edu.cn).}% <-this % stops a space
\thanks{P. Zhong is with the National Key Laboratory of Science
and Technology on ATR, College of Electrical Science and Technology,
National University of Defense Technology, Changsha 410073, China (e-mail:
zhongping@nudt.edu.cn).}
% <-this % stops a space
}


\markboth{IEEE LATEX,~Vol X, 2023}%
{Shell \MakeLowercase{\textit{et al.}}: Bare Demo of IEEEtran.cls for IEEE Journals}



\maketitle

\begin{abstract}
%To improve the performance of hyperspectral image classification, deep learning methods which can extract high level features have been applied in the image processing tasks. However, prior deep learning methods mainly train the deep models under data-driven mode while ignore the intrinsic structure within the hyperspectral images.
%In this work, we develop a novel multi-statistical model driven deep ensemble method based on convolutional neural network for hyperspectral image classification.
%First, we model the classes of the hyperspectral image with multi-statistical model. It should be noted that each class is modelled with several independent statistical model. Second, using the statistical discrimination, we assign the training samples in each training batch to different statistical model and the deep model can be trained with the cross entropy loss contructed by the gaussian probability of each sample. Finally, to balance different classes, we fused the distributions of each class for further inference. Experiments have been conducted over three real-world hyperspectral image data sets and the comparison results with other state-of-the-art methods have shown the superiority of the proposed method.

The dissection of hyperspectral images into intrinsic components through hyperspectral intrinsic image decomposition (HIID) enhances the interpretability of hyperspectral data, providing a foundation for more accurate classification outcomes. However, the classification performance of HIID is constrained by the model's representational ability. To address this limitation, this study rethinks hyperspectral intrinsic image decomposition for classification tasks by introducing deep feature embedding. The proposed framework, HyperDID, incorporates the Environmental Feature Module (EFM) and Categorical Feature Module (CFM) to extract intrinsic features. Additionally, a Feature Discrimination Module (FDM) is introduced to separate environment-related and category-related features. Experimental results across three commonly used datasets validate the effectiveness of HyperDID in improving hyperspectral image classification performance. This novel approach holds promise for advancing the capabilities of hyperspectral image analysis by leveraging deep feature embedding principles.
The implementation of the proposed method could be accessed soon at \url{https://github.com/shendu-sw/HyperDID} for the sake of reproducibility.

\end{abstract}


% \fi
\IEEEpeerreviewmaketitle

\section{Introduction}

Hyperspectral images, renowned for their ability to capture rich spectral information across a broad range of bands, offer a nuanced perspective on surface features \cite{dong2022weighted}. They play a crucial role in providing detailed and comprehensive insights into Earth's surface \cite{elmanawy2022hsi}, supporting applications such as target detection \cite{wang2022spectral, jiao2023triplet}, anomaly detection \cite{qu2020anomaly}, and land cover classification \cite{gong_manifold, b2}. In the context of land cover classification, the primary objective is to categorize pixels in hyperspectral images into predefined classes \cite{b1}, contributing to a detailed understanding of land cover, vegetation, and various surface features \cite{petropoulos2012support}. This technology is pivotal in diverse applications, including environmental monitoring \cite{zhang2012application}, agricultural management \cite{lu2020recent}, and urban planning\cite{roessner2001automated}.
However, the complexity of hyperspectral data, stemming from the high dimensionality due to numerous spectral bands, presents challenges in feature extraction and data representation. Spectral variability within and between classes further complicates the classification task, as does the presence of mixed pixels, where a single pixel may contain contributions from multiple land cover types. Feature extraction has emerged as an effective strategy to enhance the accuracy of hyperspectral image (HSI) classification \cite{kang2013feature}. Researchers have diligently developed various feature extraction methods that fully consider the spatial and spectral information inherent in HSIs \cite{he2017recent, sun2014supervised}. Additionally, extensive research has been conducted on spectral unmixing \cite{guo2021improving} and subpixel mapping methods \cite{lu2017subpixel}, further contributing to the advancement of land cover classification in hyperspectral imagery.

Hyperspectral intrinsic image decomposition (HIID), as an effective feature representation method, has gained increasing prominence in the realm of hyperspectral image processing \cite{jin2017superpixel, xie2023hyperspectral}. This technique aims to decompose hyperspectral images into intrinsic components, thereby revealing the underlying spectral characteristics of the observed scene. By isolating intrinsic features such as environment-related and category-related information, HIID enhances the interpretability of hyperspectral data, offering a more intricate understanding of spectral characteristics and ultimately leading to improved discrimination of distinct surface materials.
HIID proves particularly valuable in addressing inherent complexities associated with mixed pixels and spectral variability in hyperspectral data. It effectively mitigates spectral uncertainty, separates mixed pixel information, and extracts valuable insights for a nuanced understanding of hyperspectral content. Kang et al. \cite{kang2015intrinsic} were pioneers in applying intrinsic image decomposition for feature extraction in hyperspectral images, showcasing the method's effectiveness through experimental validation.
In subsequent research, Jin et al. \cite{jin2021intrinsic} extended the HIID framework by incorporating digital surface model (DSM) cues to effectively recover hyperspectral reflectance and environmental illumination. Similarly, Gu et al. \cite{gu2022hyperspectral} augmented HIID by integrating spatial information from high-resolution (HR) panchromatic (PAN) images, enhancing spatial details in the intrinsic component.
Despite these advancements utilizing HIID for feature extraction and enhancing classification performance, the overall efficacy is limited by the model's representational capacity. Ongoing efforts focus on overcoming these limitations and further advancing the capabilities of HIID in hyperspectral image analysis.

In the realm of hyperspectral image classification, leveraging general machine learning methods is instrumental for the extraction of meaningful features from high-dimensional spectral data. These methods encompass a diverse range, including Support Vector Machines (SVM)\cite{guo2008customizing}, Random Forests \cite{xia2017random}, Decision Trees \cite{kuching2007performance}, k-Nearest Neighbors (KNN) \cite{ma2010local}, and Ensemble Methods \cite{xia2015random}. They play a crucial role in categorizing pixels within hyperspectral images, aiming to provide a detailed understanding of land cover and enable various applications. Support Vector Machines (SVM), for instance, employ hyperplanes to effectively separate classes based on spectral features. Meanwhile, Random Forests and Decision Trees utilize decision-making processes informed by the spectral characteristics of the data. k-Nearest Neighbors (KNN) relies on the similarity of spectral signatures for classification, and Ensemble Methods amalgamate multiple models to bolster overall accuracy. To further enhance the efficiency and performance of these machine learning algorithms, techniques such as dimensionality reduction \cite{harsanyi1994hyperspectral}, feature selection \cite{kuo2013kernel}, and normalization \cite{wang2023dynamic} are routinely applied, contributing to the optimization of hyperspectral image classification by improving the extraction of pertinent features. However, it's important to note that these methods are generally considered ``shallow'' due to their limited layer depth, which may constrain their ability to capture intricate patterns and spectral information embedded in the data.

Owing to the deep architecture and the abundant network parameters, deep learning methods have demonstrated significant efficacy in the field of hyperspectral image classification  \cite{b1}.
The increased depth and richness of network parameters empower these models with a potent expressive capability, facilitating the capture of intricate patterns and detailed spectral information present in hyperspectral images \cite{lee2017going}. Consequently, this has resulted in substantial enhancements in classification accuracy and the ability to discern complex features within the data. Deep learning methods have thereby emerged as a prominent and impactful approach, driving advancements in the state-of-the-art for hyperspectral image classification \cite{li2019deep}.
Generally, the backbone networks for hyperspectral image classification can be divided into four categories: recurrent neural networks (RNNs), convolutional neural networks (CNNs), graph neural networks (GNNs), and Transformers. 
Among these backbones, RNNs allow the network to retain information from previous bands, enabling the capture of nuanced spectral patterns and temporal dynamics \cite{hang2019cascaded, mou2017deep}. GNNs treat individual pixels or spectral bands as nodes in a graph and leveraging edge connections to represent spatial dependencies, enhancing the model's ability to comprehend the complex spectral variations and spatial patterns inherent \cite{ding2022multi, hong2020graph}. Transformers enable the modeling of long-range dependencies within hyperspectral data \cite{hong2021spectralformer, zou2022lessformer}. While CNNs employ convolutional layers to automatically learn hierarchical representations, allowing for capturing intricate spectral patterns and spatial dependencies present in hyperspectral imagery \cite{roy2019hybridsn, paoletti2018new}. 

Due to the great potential of deep learning to extract intricate and high-level information from the image, this work attempt to rethink the HIID based on deep learning models to enhance the effectiveness of HIID. Through leveraging the representation power of neural networks for better feature extraction, we can obtain improved separation of intrinsic components in hyperspectral imagery.
Considering the data-driven characteristics of deep learning models, the key is how to construct the training mechanism to decompose the environment-related and category-related features. Our prior work \cite{gong2023deep} utilizes the adversarial learning methods, which can significantly improve the classification performance. However, it is worth noting that the training process encountered challenges in terms of stability. Besides, the performance is sensitive to the chosen of hyperparameters.

This work would exploit the advantages of deep feature embedding to enhance hyperspectral image classification by constructing the environmental feature module, categorical feature module, and feature discrimination module.  Deep feature embedding, known for enlarging inter-class variance and reducing intra-class variance \cite{wang2021contrastive}, is a promising approach in improving classification model performance \cite{hou2021hyperspectral, cao2021contrastnet}. Leveraging neural networks, deep feature embedding learns meaningful representations by emphasizing similarities and differences in a latent space. In hyperspectral image classification, this facilitates the extraction of discriminative features from complex spectral information, contributing to enhanced classification accuracy. The inherent capability of deep feature embedding to capture intricate patterns within high-dimensional data further strengthens its effectiveness \cite{liu2023refined, huang20223}. By emphasizing the contrast between positive and negative pairs, it enhances model robustness and generalization, making it well-suited for addressing challenges in hyperspectral data, such as mixed pixels and spectral variability. This work attempt to leverage deep feature embedding methods for the training of deep models to decompose features from hyperspectral images, obtaining discriminative environmental and categorical features.

Building upon the advantages of deep models and hyperspectral intrinsic image decomposition (HIID), this study revisits HIID within the context of deep models for hyperspectral image classification, harnessing the advantages of deep feature embedding. The proposed framework, HyperDID, accomplishes the extraction of environment-related and category-related features through the Environmental Feature Module (EFM) and Categorical Feature Module (CFM). This integration contributes to high-performance hyperspectral image classification. Additionally, the incorporation of the Feature Discrimination Module (FDM) can discern and discriminate between the distinctive characteristics associated with the environment and specific categories and effectively separate the intrinsic features. In summary, the contributions of this work can be outlined as follows.
\begin{itemize}
\item This work rethinks the hyperspectral intrinsic image decomposition within the context of deep models and develops a novel framework, called HyperDID, in order to decompose the environment-related and category-related features from the image.
\item This work develops two effective modules based on deep feature embedding, i.e. the environmental feature module (EFM) and categorical feature module (CFM), to learn the corresponding intrinsic features separately.
\item To further separate and recognize the environmental and categorical features, this work develops the feature discrimination module (FDM) discriminate different features and aids for the learning of the deep model.
\end{itemize}
Extensive experiments over three real-world and challenging datasets have demonstrated that the proposed HyperDID method can extract spatial-spectral features more effectively from the image, thereby yielding higher classification performance.

The rest of this paper is arranged as follows. Section \ref{sec:proposed} introduces the proposed HyperDID method to capture discriminative environment-related and category-related features for hyperspectral image classification. Experiments are conducted over three real-world hyperspectral image datasets to validate the effectiveness of the proposed method in Section \ref{sec:experiments}. Finally, we conclude this paper with some discussions in Section \ref{sec:conclusions}.

%yielding a high classification performance on multispectral images

\section{Proposed Method}\label{sec:proposed}

In this work, our aim is to decompose the environment-related and category-related components from a given hyperspectral image under the proposed  HyperDID framework.
For convenience, here we denote $X=\{{\bf x}_1, {\bf x}_2, \cdots, {\bf x}_N\}$ as the set of training samples from hyperspectral image, where $N$ is the number of the samples, and $y_i (i=1,2,\cdots,N)$ as the corresponding label of ${\bf x}_i$, where $y_i\in \Lambda=\{1,2,\cdots, K\}$. $K$ is the number of land cover classes in the hyperspectral image.

\subsection{Hyperspectral Intrinsic Image Decomposition}

Constructing an appropriate physical model proves to be instrumental in effectively discriminating and identifying various targets within hyperspectral images. By developing a suitable physical model, one can enhance the capability to differentiate between distinct spectral signatures associated with different objects or materials present in hyperspectral data. This approach involves leveraging domain-specific knowledge and understanding the physical principles governing the interaction of light with surfaces to create a model that accurately represents the spectral characteristics of diverse targets. The utilization of a well-designed physical model contributes to improved target discrimination and recognition, thus enhancing the overall efficacy of hyperspectral image classification.

The intrinsic information coupling model is one of such physical model to represent the hyperspectral image which refers to a system or algorithm that effectively couples and integrates intrinsic features from hyperspectral images.
This coupling of intrinsic information may involve techniques that distinguish between environment-related and category-related features, contributing to a more nuanced understanding of the spectral characteristics in the observed scene. 

As for a red-green-blue (RGB) images, the image can be typically characterized as reflectance and shading and the intrinsic images decomposition
problem can be described by
\begin{equation}
I = R \circ S
\end{equation}
where $I$ denotes the observed image,
$R$ represents the reflectance component, and $S$ stands for the shading component. Generally, $R$ represents the inherent color or texture of the scene, while $S$ accounts for variations in illumination or lighting conditions. The goal of this decomposition is to disentangle the effects of illumination and surface properties, providing a more intrinsic and scene-independent representation of the underlying scene content.

Unlike RGB images, hyperspectral images are typically acquired using passive imaging sensors designed to capture energy reflected from solar radiation. This results in pixel values across different spectral bands undergoing non-proportional changes due to variations in sensitivity to scene radiance. Consequently, the shading component of hyperspectral images has varying effects on each wavelength. Therefore, the hyperspectral intrinsic image decomposition can be formulated as
\begin{equation}
I(\lambda)=R(\lambda)\circ S(\lambda)
\end{equation}
where $\lambda$ stands for the wavelength.
$R(\lambda)$ and $\lambda$ denotes the reflectance and shading component, respectively. 
Generally, $R$ determines the spectral signatures in hyperspectral images and $S$ is the features influenced by the environmental factors.

Following the above-mentioned assumptions, we will introduce the detailed framework of proposed HyperDID to learn the deep model which decompose the hyperspectral image into the category-related features $R(\lambda)$ and the environmental-related features $S(\lambda)$.

\subsection{Overall Framework of HyperDID}\label{subsec:overall}

\begin{figure*}[t]
\centering
\includegraphics[width=0.9\linewidth]{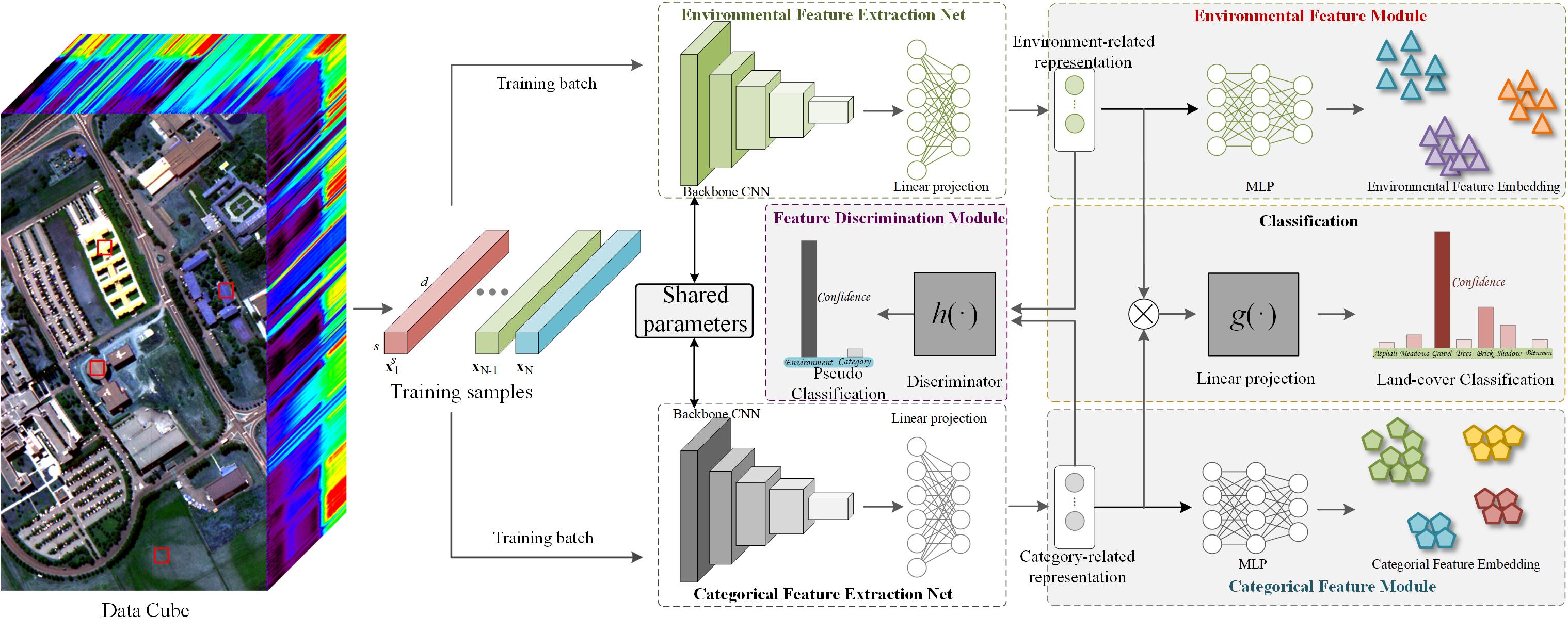}
   \caption{Overall framework of Hyperspectral Intrinsic Image Decomposition with Deep Feature Embedding (HyperDID) method for hyperspectral image classification. }
\label{fig:flowchart}
\vspace{-0.5cm}
\end{figure*}

To harness the robust representational ability of deep models for hyperspectral intrinsic image decomposition (HIID), we introduce HyperDID, a novel framework that reimagines HIID through the lens of deep feature embedding. 
To this end, we develop three key modules, i.e., Environmental Feature Module (EFM), Categorical Feature Module (CFM), Feature Discrimination Module (FDM), and take advantage of these modules to construct the feature extract network. 
The overall architecture of HyperDID is illustrated in Fig. \ref{fig:flowchart}. 

Broadly, HyperDID employs a Convolutional Neural Network (CNN) as the backbone to extract features from the hyperspectral image.
As depicted in Fig. \ref{fig:flowchart}, the CNN backbone is employed for extracting discriminative features, followed by two parallel Multi-Layer Perceptrons (MLPs). These MLPs work in tandem to extract environment-related and category-related features, respectively. 
Denote $f_1(\cdot)$, $f_2(\cdot)$ as the representation function of Environmental Feature Extraction Net and Categorical Feature Extraction Net, then the aim of HyperDID is to find the optimal $f_1(\cdot)$, $f_2(\cdot)$, by solving the following optimization problem:
\begin{equation}\label{eq:01}
\min\limits_{f_1, f_2}\sum_{i=1}^N C_1(g(f_1({\bf x}_i)\circ f_2({\bf x}_i)), y_i)
\end{equation}
where $g(\cdot):\mathbb{R}^d\rightarrow \mathbb{R}^K$ is the mapping function from the features to classification probabilities and $C_1(\cdot)$ denotes a classification loss function. $d$ stands for the dimension of the learned feature from the image.
$f_1({\bf x}_i)$ and $f_2({\bf x}_i)$ represent the environmental-related features $S(\lambda)$ and the category-related features $R(\lambda)$, respectively.

To solve the optimization \ref{eq:01}, the proposed modules in HyperDID are then followed to formulate the feature extraction network, where the EFM is designed to learn a subnet specifically for environmental features, the CFM is  tailored to learn a subnet dedicated to category-related features, and the FDM discriminates the categorical and environmental features.

In the subsequent sections, we provide a detailed introduction to each of the developed modules based on deep feature embedding, elucidating their roles and functionalities within the HyperDID framework. This comprehensive approach aims to enhance the capabilities of HIID and push the boundaries of hyperspectral image decomposition for improved classification performance.

\subsection{Environmental Feature Module}

The goal of the Environmental Feature Module (EFM) is to effectively capture and represent the environment-related features within hyperspectral images. By leveraging deep contrastive learning, EFM aims to discern and extract intrinsic information associated with the environmental components, such as illumination, shading, and other factors that contribute to the overall spectral characteristics of the scene. Through this process, EFM contributes to the disentanglement of environment and category-related features, enhancing the model's ability to discern meaningful information for hyperspectral image classification. The environmental features captured by EFM play a crucial role in improving the overall interpretability and discrimination capabilities of the HyperDID framework.

\subsubsection{Construction of Environmental Pseudo Classes}

First, specific training samples are utilized to construct environmental pseudo classes in an unsupervised manner. This entails clustering or grouping pixels based on their spectral characteristics, creating distinct classes that represent various environmental components. The unsupervised construction of environmental pseudo classes enables the model to autonomously discern patterns and groupings within the hyperspectral data, capturing the inherent complexity and diversity of environmental features. This preparatory step lays the foundation for subsequent training and learning processes within the HyperDID framework, facilitating the extraction of meaningful environment-related information for enhanced hyperspectral image classification.
By employing clustering techniques, we aim to group pixels with similar spectral characteristics into distinct environmental pseudo classes, effectively categorizing them based on shared features. This clustering process enables the identification and differentiation of various environmental components within the hyperspectral data, contributing to the establishment of robust environmental pseudo classes. The utilization of clustering methods ensures an unsupervised confirmation of environmental categories, providing a foundational step for subsequent stages in the HyperDID framework.

Assume that there are $\Lambda$ environmental pseudo classes in the hyperspectral image. Denote $P_s(s=1,2,\cdots, \Lambda)$ as the environmental centers of different pseudo classes. Iteratively, we calculate the centers of the $\Lambda$ classes, $P_s(s=1,2,\cdots, \Lambda)$, solving the following optimization problem:
\begin{equation}\label{eq:003}
\min\limits_{P_1, P_2, \cdots, P_\Lambda} \sum_{i=1}^N\sum_{s=1}^\Lambda I(s=\arg\min\limits_{1,2,\cdots,\Lambda}\|{\bf x}_i-P_s\|^2)\|{\bf x}_i-P_s\|^2
\end{equation}
where $I(condition)$ denotes the indicative function where $I(\cdot)=1$ if condition is true and $I(\cdot)=0$ otherwise.

We can construct the environmental pseudo classes following Eq. \ref{eq:003}. Based on the cluster centers obtained through clustering, environmental pseudo-labels are assigned to all hyperspectral image samples using the Euclidean distance and the nearest-neighbor principle. This process involves calculating the Euclidean distance between each pixel's spectral signature and the cluster centers. Subsequently, each pixel is assigned an environmental pseudo-label based on the principle of proximity, where the label corresponds to the nearest cluster center. 
Therefore, denote $z_i$ as the corresponding environmental pseudo label of ${\bf x}_i$, then we can calculate the environmental pseudo label $z_i$ by
\begin{equation}\label{eq:004}
z_i=\arg\min\limits_{s=1,2,\cdots,\Lambda} \|{\bf x}_i-P_s\|^2
\end{equation}

\subsubsection{Environmental Feature Embedding}

The aim of environmental feature embedding is to enlarge the inter-class variance and decrease the intra-class variance of the obtained environmental features.

Given a training batch $B=\{{\bf x}_{b_1},{\bf x}_{b_2},\cdots, {\bf x}_{b_m}\}$ in the image where $m$ is size of training batch. From Eq. \ref{eq:004}, we can obtain the corresponding environmental pseudo class ${z_{b_1}, z_{b_2}, \cdots, z_{b_m}}$.

A projection $p_1(\cdot)$ maps the environmental representation into a vector representation which is more suitable for contrastive learning. We implement this projection head $p_1(\cdot)$ as a nonlinear multiple-layer perceptron (MLP). Such project module is proven improtent in improving the representation quality \cite{chen2020simple}.

This work uses the cosine embedding loss for the training of environmental feature net.
Given two samples ${\bf x}_{b_i}$, ${\bf x}_{b_j}$ in a given batch, the loss $l_{ij}$ can be calculated as
\begin{equation}
l_{ij}=\left\{
\begin{array}{lr}
1-\cos(p_1(f_1({\bf x}_{b_i})), p_1(f_1({\bf x}_{b_j}))), &\text{if} \ z_{b_i}=z_{b_j}\\
\max(0,\cos(p_1(f_1({\bf x}_{b_i})), p_1(f_1({\bf x}_{b_j})))-\delta), &\text{if} \ z_{b_i}\neq z_{b_j}\\
\end{array}
\right.
\end{equation}
where $\delta$ stands for the margin which we set to 0 in the experiments.

Therefore, 
for the given batch $B$, the environmental feature embedding loss can be constructed as
\begin{equation}
L_e = \sum_{i=1}^{m}\sum_{j=1}^m l_{ij}.
\end{equation}
Under $L_e$, the obtained environmental feature can be more discriminative to recognize different environmental pattern.

\subsection{Categorical Feature Module}

The objective of the Categorical Feature Module (CFM) within the HyperDID framework is to identify and categorize the ``category-related features'' present in the hyperspectral image. In this context, ``category-related features'' refer to the spectral signatures associated with different land cover or material classes. The CFM operates in conjunction with the Environmental Feature Module (EFM) to jointly capture both environment and category-related information, ensuring a comprehensive understanding of the intrinsic components within the hyperspectral data.

Similarly, a projection $p_2(\cdot)$ maps the environmental representation into a vector representation which is more suitable for the learning of categorical features. 

Assuming the training batch $B$ as the former subsection, this work also uses the cosine embedding loss for the training of categorical feature net. Given two samples ${\bf x}_{b_i}$, ${\bf x}_{b_j}$ in a given batch with ${y}_{b_i}$, ${y}_{b_j}$ as the corresponding labels, the loss $l^c_{ij}$ can be calculated as
\begin{equation}
l^c_{ij}=\left\{
\begin{array}{lr}
1-\cos(p_2(f_2({\bf x}_{b_i})), p_2(f_2({\bf x}_{b_j}))), &\text{if} \ y_{b_i}=y_{b_j}\\
\max(0,\cos(p_2(f_2({\bf x}_{b_i})), p_2(f_2({\bf x}_{b_j})))-\delta), &\text{if} \ y_{b_i}\neq y_{b_j}\\
\end{array}
\right.
\end{equation}
For the given batch $B$, the environmental feature embedding loss can be constructed as
\begin{equation}
L_c = \sum_{i=1}^{m}\sum_{j=1}^m l^c_{ij}.
\end{equation}
Under $L_c$, the obtained categorical feature can be more discriminative to recognize different categories in the hyperspectral image.

\subsection{Feature Discrimination Module}

The Feature Discrimination Module (FDM) in the HyperDID framework serves the purpose of enhancing the discriminative ability between environment-related and category-related features extracted by the Environmental Feature Module (EFM) and the Categorical Feature Module (CFM), respectively. The primary goal is to refine the separation of intrinsic components, ensuring a more accurate decomposition of hyperspectral images.

The FDM achieves feature discrimination through mechanisms that emphasize the differences in the learned representations of environment-related and category-related features.
This work constructs a feature discrimination classifier to discriminate the categorical and environmental features.

First, a feature discriminator $h(\cdot)$ is designed to map the features, e.g. environment-related representation and category-related representation to classification probabilities concerning the types of the features, e.g. the Environmental feature or categorical feature.

Following the former subsections, we also formulate the classification loss under the training batch $B$.
Specifically, $f_1({\bf x}_{b_i})(i=1,2,\cdots,m)$ are the environment-related features, which are labelled as 0 and $f_2({\bf x}_{b_i})(i=1,2,\cdots,m)$ are the category-related features, which are labelled as 1.

Then, the feature discrimination loss can be formulated as
\begin{equation}
L_d = \sum_{i=1}^m(C_2(h(f_1({\bf x}_{b_i})), 0) + C_2(h(f_2({\bf x}_{b_i})), 1))
\end{equation}
where $C_2$ denotes a classification loss function. In this work, we use cross entropy loss for $C_2(\cdot)$, which is given as
\begin{equation}
\begin{aligned}
C_2(h(f_1({\bf x}_{b_i})), 0)=-\log(h(f_1({\bf x}_{b_i}))^Te_0)   \\
C_2(h(f_2({\bf x}_{b_i})), 1)=-\log(h(f_2({\bf x}_{b_i}))^Te_1)
\end{aligned}
\end{equation}
where $e_0\in \mathbb{R}^2$ and $e_1\in \mathbb{R}^2$ represent the standard basis vector.

Under $L_d$, the categorical and environmental features are subjected to a learning process that encourages distinctiveness between different categories. This distinctiveness is crucial for decomposing the learned features from the hyperspectral image.

\subsection{HyperDID for Classification}

The CFM, EFM, and FDM module are essential to obtain a good hyperspectral intrinsic image decomposition.
However, to solve the optimization in Eq. \ref{eq:01}, a classification loss is also required. Under the training batch $B$, the classification loss $L_0$ can be formulated as
\begin{equation}
L_0=\sum_{i=1}^m C_1(g(f_1({\bf x}_{b_i})\circ f_2({\bf x}_{b_i})), y_{b_i})
\end{equation}
As subsection \ref{subsec:overall} shows, $C_1$ denotes a classification loss function which is also formulated by cross entropy loss,
\begin{equation}
C_1(g(f_1({\bf x}_{b_i})\circ f_2({\bf x}_{b_i})), y_{b_i})=-\sum_{j=1}^K\delta_{jy_i}\log(g(f_1({\bf x}_{b_i})\circ f_2({\bf x}_{b_i}))^Te_j)
\end{equation}
where $e_j\in\mathbb{R}^K$ stands for the standard basis vector.

Therefore, the overall training loss of the HyperDID can be formulated as
\begin{equation}\label{eq:loss}
L=L_0+\alpha L_e+\beta L_c+\gamma L_d
\end{equation}
For convenience, all the hyperparameter of $\alpha$, $\beta$ and $\gamma$ are set to 1.
Trained with Eq. \ref{eq:loss}, we can obtain discriminative features from the hyperspectral image and recognize different samples with the obtained features.

\section{Experimental Results}\label{sec:experiments}
\subsection{Data Description}
In this section, the proposed HyperDID method is evaluated over three real-world hyperspectral image data, namely, the Pavia University data \cite{pavia}, the Indian Pines data \cite{pavia}, and the Houston 2013 data  \cite{houston2013}.

{\bf Pavia University (PU) data} was collected by the reflective optics system imaging spectrometer (ROSIS-3) sensor ranging from 0.43 to 0.86 $\mu m$ with a spatial resolution of 1.3$m$ per pixel. The data contains $610\times 340$ pixels with 115 bands where 12 bands have removed due to noise and absorption and the remaining 103 channels are used. It contains 9 different classes, representing various land cover categories such as asphalt, trees, and buildings and 42776 samples from different categories have been labeled for experiments. Table. \ref{table:pavia} has presented the detailed training and testing samples of the data.

{\bf Indian Pines (IP) data} was gathered by the 224-band Airborne Visible/Infrared Imaging Spectrometer (AVIRIS) sensor ranging from 0.4 to 2.5 $\mu m$ over the over the Indian Pines test site in Indiana, USA. The high spectral resolution of the Indian Pines data enables detailed characterization of land cover and vegetation types. With 16 different classes representing various ground cover categories such as crops, trees, and soil (see Table \ref{table:indian} for details), this dataset serves as a benchmark for evaluating and testing algorithms in the experiments. The data consists of $145\times 145$ pixels with 200 clean spectral bands, and 10366 labeled samples are selected for experiments as shown in the Table \ref{table:indian}.

{\bf Houston 2013 (HS) data} was acquired over the University of Houston campus and the neighboring urban area with a resolution of 2.5 m/pixel, and provided as part of the 2013 IEEE Geoscience and remote Sensing Society data fusion contest.
It consists of $349\times 1905$ pixels of which a total of 15011 labeled samples divided into 15 classes have been chosen for experiments. Each pixel denotes a sample and contains 144 spectral bands ranging from 0.38 to 1.05 $\mu m$. The experimental details of the dataset are listed in the Table \ref{table:houston2013}.

\subsection{Experimental Setups}
\subsubsection{Evaluation Metrics}
The experiments assess the performance of all methods using three widely adopted metrics: overall accuracy (OA), average accuracy (AA), and the Kappa coefficient $\kappa$. Additionally, a complementary metric is incorporated, detailing the classification accuracy for each individual class. This comprehensive evaluation framework ensures a thorough and nuanced analysis of the methods' effectiveness across various dimensions of performance.

\subsubsection{Implementation Details}
Pytorch 1.9.1, Cuda 11.2 is chosen as the deep learning framework to implement the proposed method  \cite{paszke2019pytorch}.
By default, the learning rate, epoch iteration, and training batch are set to 0.01, 500, and 64, respectively and the dimension of extracted environment-related and category-related features is set to 128.
The structures of MLPs for non-linear projection in deep feature embedding are set as `$128-64-64$' and $5\times 5$ neighbors are used for classification.
The code would be publicly available soon at \url{https://github.com/shendu-sw/HyperDID}.

\subsubsection{Baseline Methods}\label{subsubsec:baseline}

In the context of this study, an extensive repertoire of state-of-the-art methodologies has been purposefully curated as baseline benchmarks. This meticulous selection spans cutting-edge Convolutional Neural Networks (CNNs), including 3-D CNN \cite{hamida20183}, PResNet \cite{paoletti2018deep}, and HybridSN \cite{roy2019hybridsn}, which exemplify sophisticated techniques in hyperspectral image processing. The integration of Recurrent Neural Networks (RNNs), particularly the RNN architecture, is geared towards capturing sequential dependencies inherent in hyperspectral data  and this work selects the RNN model in \cite{hang2019cascaded} as a representative. Additionally, Graph Convolutional Networks (GCNs), embodied by miniGCN \cite{hong2020graph}, underscore the significance of graph-based learning methodologies in this domain. The inclusion of Transformers, featuring models like ViT \cite{dosovitskiy2020image}, SpectralFormer \cite{hong2021spectralformer}, and SSFTTNet \cite{sun2022spectral}, highlights the transformative capabilities of attention mechanisms.

This comprehensive array of carefully chosen state-of-the-art methods serves as a robust foundation for conducting a thorough comparison across diverse neural network architectures. It not only accentuates the effectiveness and efficiency of the proposed HyperDID method for hyperspectral image classification but also ensures a well-rounded evaluation. Beyond neural networks, the study incorporates the Support Vector Machine (SVM) with a radial basis function kernel, enriching the comparison framework to gain nuanced insights into the strengths and limitations of various techniques in hyperspectral image analysis.
%\begin{table}[t]
%\centering
%\caption{Architecture of the CNN}
%\begin{tabular}{|c|c|c|c|c|}
%  \hline
%  % after \\: \hline or \cline{col1-col2} \cline{col3-col4} ...
%  {\bf spatial size} & {\bf conv1} & {\bf conv2} & {\bf FC1} & {\bf FC2} \\
%  \hline
%  $3\times 3$, $5\times 5$ & $1\times 1$, 1 & $1\times 1$, 1 & 400 & 100 \\
%  $7\times 7$, $9\times 9$ & $3\times 3$, 2 & $1\times 1$, 1 & 400 & 100 \\
%  $11\times 11$, $13\times 13$ & $3\times 3$, 2 & $3\times 3$, 2 & 400 & 100 \\
%  \hline
%\end{tabular}
%\end{table}

\begin{table}[t]
\begin{center}
\caption{Number of training and testing samples in Pavia University data.}
\label{table:pavia}
%\vspace{-1ex}
\begin{tabular}{ c | c c c c }
\toprule[1pt]
{ Class}     &  { Class Name} & Color &  { Training}&  { Testing} \\
\hline\hline
C1   &  Asphalt                  &\begin{minipage}[b]{0.08\columnwidth}
		\raisebox{-.45\height}{\includegraphics[width=\linewidth]{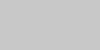}}
	\end{minipage}  & 548  & 6304  \\
C2   &  Meadows                  & \begin{minipage}[b]{0.08\columnwidth}
		\raisebox{-.45\height}{\includegraphics[width=\linewidth]{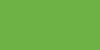}}
	\end{minipage} & 540  & 18146 \\
C3   &  Gravel                   & \begin{minipage}[b]{0.08\columnwidth}
		\raisebox{-.45\height}{\includegraphics[width=\linewidth]{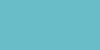}}
	\end{minipage} & 392  & 1815   \\
C4   &  Trees                    & \begin{minipage}[b]{0.08\columnwidth}
		\raisebox{-.45\height}{\includegraphics[width=\linewidth]{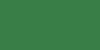}}
	\end{minipage} & 524  & 2912   \\
C5   &  Metal sheet              & \begin{minipage}[b]{0.08\columnwidth}
		\raisebox{-.45\height}{\includegraphics[width=\linewidth]{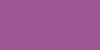}}
	\end{minipage} & 265  & 1113  \\
C6   &  Bare soil                & \begin{minipage}[b]{0.08\columnwidth}
		\raisebox{-.45\height}{\includegraphics[width=\linewidth]{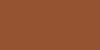}}
	\end{minipage} & 532  & 4572  \\
C7   &  Bitumen                  & \begin{minipage}[b]{0.08\columnwidth}
		\raisebox{-.45\height}{\includegraphics[width=\linewidth]{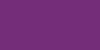}}
	\end{minipage} & 375  & 981   \\
C8   &  Brick                    & \begin{minipage}[b]{0.08\columnwidth}
		\raisebox{-.45\height}{\includegraphics[width=\linewidth]{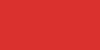}}
	\end{minipage} & 514  & 3364   \\
C9   &  Shadow                   & \begin{minipage}[b]{0.08\columnwidth}
		\raisebox{-.45\height}{\includegraphics[width=\linewidth]{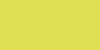}}
	\end{minipage} & 231  & 795  \\

 \hline\hline
Total     &         &  & 3921 & 40002  \\
\bottomrule[1pt]
\end{tabular}
\end{center}
\vspace{-0.5cm}
\end{table}

\begin{table}[t]
\begin{center}
\caption{Number of training and testing samples in Indian Pines data.}
\label{table:indian}
%\vspace{-1ex}
\begin{tabular}{ c | c c c c }
\toprule[1pt]
{Class}     &  {Class Name} & Color &  {Training}&  {Testing}  \\
\hline\hline
C1   &  Corn-notill                    &\begin{minipage}[b]{0.08\columnwidth}
		\raisebox{-.45\height}{\includegraphics[width=\linewidth]{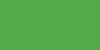}}
	\end{minipage} & 50  & 1384   \\
C2   &  Corn-mintill                   &\begin{minipage}[b]{0.08\columnwidth}
		\raisebox{-.45\height}{\includegraphics[width=\linewidth]{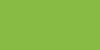}}
	\end{minipage} & 50  & 784  \\
C3   &  Corn                           &\begin{minipage}[b]{0.08\columnwidth}
		\raisebox{-.45\height}{\includegraphics[width=\linewidth]{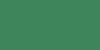}}
	\end{minipage} & 50  & 184   \\
C4   &  Grass-pasture                  &\begin{minipage}[b]{0.08\columnwidth}
		\raisebox{-.45\height}{\includegraphics[width=\linewidth]{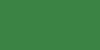}}
	\end{minipage} & 50  & 447   \\
C5   &  Grass-trees                    &\begin{minipage}[b]{0.08\columnwidth}
		\raisebox{-.45\height}{\includegraphics[width=\linewidth]{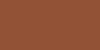}}
	\end{minipage} & 50  & 697  \\
C6   &  Hay-windrowed                  &\begin{minipage}[b]{0.08\columnwidth}
		\raisebox{-.45\height}{\includegraphics[width=\linewidth]{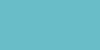}}
	\end{minipage} & 50  & 439   \\
C7   &  Soybean-notill                 &\begin{minipage}[b]{0.08\columnwidth}
		\raisebox{-.45\height}{\includegraphics[width=\linewidth]{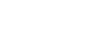}}
	\end{minipage} & 50  & 918   \\
C8   &  Soybean-mintill                &\begin{minipage}[b]{0.08\columnwidth}
		\raisebox{-.45\height}{\includegraphics[width=\linewidth]{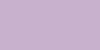}}
	\end{minipage} & 50  & 2418   \\
C9   &  Soybean-clean                  &\begin{minipage}[b]{0.08\columnwidth}
		\raisebox{-.45\height}{\includegraphics[width=\linewidth]{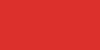}}
	\end{minipage} & 50  & 564  \\
C10   & Wheat                          &\begin{minipage}[b]{0.08\columnwidth}
		\raisebox{-.45\height}{\includegraphics[width=\linewidth]{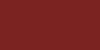}}
	\end{minipage} & 50  & 162  \\
C11   & Woods                          &\begin{minipage}[b]{0.08\columnwidth}
		\raisebox{-.45\height}{\includegraphics[width=\linewidth]{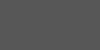}}
	\end{minipage} & 50  & 1244  \\
C12   & Buildings-Grass-Trees-Drives   &\begin{minipage}[b]{0.08\columnwidth}
		\raisebox{-.45\height}{\includegraphics[width=\linewidth]{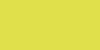}}
	\end{minipage} & 50  & 330  \\
C13   & Stone-Steel-Towers             &\begin{minipage}[b]{0.08\columnwidth}
		\raisebox{-.45\height}{\includegraphics[width=\linewidth]{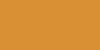}}
	\end{minipage} & 50  & 45  \\
C14   & Alfalfa                        &\begin{minipage}[b]{0.08\columnwidth}
		\raisebox{-.45\height}{\includegraphics[width=\linewidth]{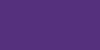}}
	\end{minipage} & 15  & 39  \\
C15   & Grass-pasture-mowed            &\begin{minipage}[b]{0.08\columnwidth}
		\raisebox{-.45\height}{\includegraphics[width=\linewidth]{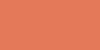}}
	\end{minipage} & 15  & 11  \\
C16   & Oats                           &\begin{minipage}[b]{0.08\columnwidth}
		\raisebox{-.45\height}{\includegraphics[width=\linewidth]{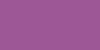}}
	\end{minipage} & 15  & 5  \\
 \hline\hline
Total     &                            & & 695 & 9671  \\
\bottomrule[1pt]
\end{tabular}
\end{center}
\vspace{-0.5cm}
\end{table}

\begin{table}[t]
\begin{center}
\caption{Number of training and testing samples in Houston2013 data.}
\label{table:houston2013}
%\vspace{-1ex}
\begin{tabular}{ c | c c c c }
\toprule[1pt]
{Class}     &  {Class Name} & Color&  {Training} &  {Testing} \\
\hline\hline
C1   &  Grass-healthy   &  \begin{minipage}[b]{0.08\columnwidth}
		\raisebox{-.45\height}{\includegraphics[width=\linewidth]{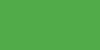}}
	\end{minipage}& 198  & 1053   \\
C2   &  Grass-stressed  &  \begin{minipage}[b]{0.08\columnwidth}
		\raisebox{-.45\height}{\includegraphics[width=\linewidth]{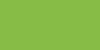}}
	\end{minipage}& 190  & 1064  \\
C3   &  Grass-synthetic &  \begin{minipage}[b]{0.08\columnwidth}
		\raisebox{-.45\height}{\includegraphics[width=\linewidth]{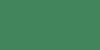}}
	\end{minipage}& 192  & 505   \\
C4   &  Tree            & \begin{minipage}[b]{0.08\columnwidth}
		\raisebox{-.45\height}{\includegraphics[width=\linewidth]{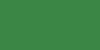}}
	\end{minipage} & 188  & 1056   \\
C5   &  Soil            &  \begin{minipage}[b]{0.08\columnwidth}
		\raisebox{-.45\height}{\includegraphics[width=\linewidth]{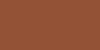}}
	\end{minipage}& 186  & 1056  \\
C6   &  Water           &  \begin{minipage}[b]{0.08\columnwidth}
		\raisebox{-.45\height}{\includegraphics[width=\linewidth]{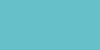}}
	\end{minipage}& 182  & 143   \\
C7   &  Residential     &  \begin{minipage}[b]{0.08\columnwidth}
		\raisebox{-.45\height}{\includegraphics[width=\linewidth]{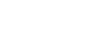}}
	\end{minipage}& 196  & 1072   \\
C8   &  Commercial      &  \begin{minipage}[b]{0.08\columnwidth}
		\raisebox{-.45\height}{\includegraphics[width=\linewidth]{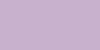}}
	\end{minipage}& 191  & 1053   \\
C9   &  Road            &  \begin{minipage}[b]{0.08\columnwidth}
		\raisebox{-.45\height}{\includegraphics[width=\linewidth]{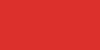}}
	\end{minipage}& 193  & 1059  \\
C10   & Highway         &  \begin{minipage}[b]{0.08\columnwidth}
		\raisebox{-.45\height}{\includegraphics[width=\linewidth]{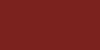}}
	\end{minipage}& 191  & 1036  \\
C11   & Railway         &  \begin{minipage}[b]{0.08\columnwidth}
		\raisebox{-.45\height}{\includegraphics[width=\linewidth]{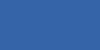}}
	\end{minipage}& 181  & 1054  \\
C12   & Parking-lot1    &  \begin{minipage}[b]{0.08\columnwidth}
		\raisebox{-.45\height}{\includegraphics[width=\linewidth]{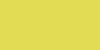}}
	\end{minipage}& 192  & 1041  \\
C13   & Parking-lot2    &  \begin{minipage}[b]{0.08\columnwidth}
		\raisebox{-.45\height}{\includegraphics[width=\linewidth]{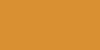}}
	\end{minipage}& 184  & 285  \\
C14   & Tennis-court    & \begin{minipage}[b]{0.08\columnwidth}
		\raisebox{-.45\height}{\includegraphics[width=\linewidth]{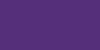}}
	\end{minipage} & 181  & 247  \\
C15   & Running-track   &  \begin{minipage}[b]{0.08\columnwidth}
		\raisebox{-.45\height}{\includegraphics[width=\linewidth]{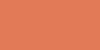}}
	\end{minipage}& 187  & 473 \\

 \hline\hline
Total     &             & & 2832 & 12197  \\
\bottomrule[1pt]
\end{tabular}
\end{center}
\vspace{-0.5cm}
\end{table}

\subsection{Evaluation of Computational Performance}

In our investigation of the computational performance of the proposed HyperDID, we employ the HybridSN as the chosen backbone CNN for feature extraction from the image. To showcase the versatility and general applicability of the proposed method, we conduct experiments on a standard machine equipped with an Intel Xeon(R) Gold 6226R CPU, 128GB RAM, and a Quadro RTX 6000 24GB GPU for evaluating classification performance. As benchmarks for comparison, we consider the training and testing costs of 3-D CNN, PResNet, HybridSN, and SpectralFormer. This comprehensive evaluation aims to provide insights into the efficiency and effectiveness of HyperDID in comparison to established baseline models. Table \ref{table:computational} presents the detailed comparison results over the three datasets.

The outcomes presented in Table \ref{table:computational} shed light on the comparative performance of the proposed HyperDID against established baseline models across three datasets. The training durations for HyperDID are notably competitive, clocking in at approximately 646.7s, 147.0s, and 400.7s for the respective datasets. These results position HyperDID as a computationally efficient alternative, performing on par with 3-D CNN and HybridSN, and surpassing PResNet and SpectralFormer in terms of training efficiency.  Moreover, the testing phase demonstrates the practicality of HyperDID, with durations of about 1.88s, 0.61s, and 0.69s for the three datasets, catering to the needs of various applications. Notably, this indicates that the proposed HyperDID achieves significant performance improvements without introducing additional time costs, reaffirming its suitability for real-world applications where both accuracy and computational efficiency are crucial considerations.

\begin{table}[t]
\begin{center}
\caption{Computational performance over different datasets.}
\label{table:computational}
%\vspace{-1ex}
\begin{tabular}{ c | c || c c c}
\toprule[1pt]
{Methods}     & {Stage} &  {PU} &  IP&  HS  \\
\hline\hline
  \multirow{2}{*}{3-D CNN}  &  Training(s)        &  471.24 & 112.7&308.9  \\
  &  Testing(s)      &  1.225 & 0.45&0.425  \\
\hline
 \multirow{2}{*}{PResNet}&  Training(s)  &  1752.4 & 364.0&961.24  \\
  &  Testing(s)        &  7.1 & 2.1&2.2  \\
\hline 
\multirow{2}{*}{HybridSN}  &  Training(s)      &  536.2 & 123.9&320.5 \\
   &  Testing(s)  &  1.7 & 0.56&0.55  \\
\hline 
\multirow{2}{*}{SpectralFormer}  & Training(s)      &  1061.2 & 232.9&621.3  \\
   &  Testing(s)  &  2.52 & 0.92&0.77  \\
\hline 
\multirow{2}{*}{AdverDecom}  &  Training(s)      &  646.7 & 147.0&400.7    \\
   &  Testing(s)  &  1.88 & 0.61&0.69   \\
%\hline
%   &  OA(\%)       & 86.30  &88.60 &88.93    \\
%  HS &  AA(\%)     & 88.34  &89.91 &90.72    \\
%   &  $\kappa$(\%) & 85.12  &87.63 &87.99    \\
\bottomrule[1pt]
\end{tabular}
\end{center}
\vspace{-0.5cm}
\end{table}

\subsection{Comparative Analysis of Models Trained with Different Hyperparameters}

In this experimental set, we systematically investigate the impact of various hyperparameters within the HyperDID framework on classification accuracies. Our exploration involves the deliberate modification of hyperparameter values to discern their influence on the overall performance of the model. Specifically, we focus on two key categories of hyperparameters: those associated with training, such as $\alpha, \beta, \gamma$, and those integral to the implementation of deep feature embedding, including the number of environmental pseudo classes. Both sets of hyperparameters are recognized for their pivotal roles in shaping the classification performance of HyperDID.

The subsequent discussion delves into a detailed examination of the effects of these hyperparameters, shedding light on their individual contributions to the overall efficacy of the model. 
This nuanced exploration of hyperparameter effects would contribute to a deeper understanding of the model's behavior and its adaptability to different configurations, thereby enhancing its versatility and performance across various scenarios.

\subsubsection{Effects of Training Hyperparameters}

The examination of training hyperparameters in HyperDID reveals crucial insights into the model's performance. Table \ref{table:training_hyper} provides a detailed breakdown of the classification performance under varying hyperparameter settings across three datasets. Specifically, $\alpha$, $\beta$, and $\gamma$ are examined to understand the tradeoffs associated with EFM, CFM, and FDM, respectively. These hyperparameters dictate the importance given to each module during training, with default values of 1 and the option to set a parameter to 0, indicating the exclusion of the corresponding module from the training process.

Upon close analysis of the results, it becomes evident that each module within HyperDID plays an indispensable role. Notably, models lacking the FDM or EFM module exhibit the lowest classification accuracies. When either the FDM or EFM module is introduced, there is a significant enhancement in the classification results compared to models without these modules, with improvements of approximately 2\%, 1\%, and 2\% OA observed across the three datasets, respectively. This underscores the synergistic contribution of all three modules to the overall effectiveness of HyperDID, highlighting the importance of carefully tuning these hyperparameters for optimal model performance.

\begin{table}[t]
\begin{center}
\caption{Classification accuracies (OA, AA, and $\kappa$) of the proposed method with different hyperparameter settings.}
\label{table:training_hyper}
%\vspace{-1ex}
\begin{tabular}{ c | c || c c c}
\toprule[1pt]
  Data & Settings   &  {OA (\%)}&  {AA (\%)}&  {$\kappa$ (\%)}   \\
\hline\hline
    \multirow{4}{*}{PU}    &  $\alpha=0$ &   92.92 & 91.79&90.42   \\
        &  $\beta=0$  &    94.01 & 93.54&91.95    \\
        &  $\gamma=0$    &  92.76 & 92.79&90.21    \\
        &  $-$   &  {\bf 94.45} & {\bf 94.36}&{\bf 92.53}    \\
\hline
    \multirow{4}{*}{IP}    &   $\alpha=0$  &  88.76 & 94.07&87.16   \\
        &  $\beta=0$   &  89.25 & 93.68&87.71    \\
        &  $\gamma=0$  &  88.51 & 93.56&86.83    \\
        & $-$  & {\bf 89.40} & {\bf 94.11} &{\bf 87.85}    \\
\hline
    \multirow{4}{*}{HS}    &   $\alpha=0$ &  87.55 & 89.98&86.48   \\
        &  $\beta=0$  &  89.73 & 91.26&88.85    \\
        &  $\gamma=0$  &  87.51 & 89.51&86.44    \\
        & $-$  &  {\bf 89.74} & {\bf 91.57} & {\bf 88.87}    \\
\bottomrule[1pt]
\end{tabular}
\end{center}
\vspace{-0.5cm}
\end{table}

\subsubsection{Effects of Number of Environmental Pseudo Classes} 

In addition to the intricate interplay between learnable parameters and training hyperparameters, the number of environmental pseudo classes emerges as a pivotal factor in shaping the success of feature embedding for environmental-related features within HyperDID. A meticulous exploration of the appropriate parameter range for this crucial aspect is essential, and our investigation extends to evaluating parameter sensitivity across three distinct datasets, as outlined in Table \ref{table:pseudo_class}. This table delineates the evolving trends of classification accuracies concerning OA, AA, and $\kappa$ as the number of environmental pseudo classes incrementally varies.

A noteworthy consensus arising from our observations is that leveraging environmental pseudo classes through clustering analysis imparts valuable prior environmental information for the learning process. This, in turn, enhances the model's ability to discern and extract features that are both environmentally and categorically relevant from the input samples. Within a discernible range, our findings reveal a robust insensitivity of the parameter to classification performance, indicating a stable operational zone. This characteristic not only underscores the efficacy of the proposed model but also highlights its potential for practical applications. The implication is that the identified parameter can be readily applied to other datasets within similar contexts, streamlining the adaptability and utility of HyperDID across diverse classification tasks.

\begin{table*}[t]
\begin{center}
\caption{Classification accuracies (OA, AA, and $\kappa$) of the proposed method with different number of environmental pseudo classes.}
\label{table:pseudo_class}
%\vspace{-1ex}
\begin{tabular}{ c | c || c c c c c c c c c c c c c}
\toprule[1pt]
\multirow{2}{*}{Data}     &  \multirow{2}{*}{Metrics} &  \multicolumn{12}{c}{Number of Environmental Pseudo Classes}  \\
\cline{3-14}
     &   &  { 1}&  { 2}&  { 3} &  {4} & {5} & {6} & {7} & {8} & {9} & 10 & 20 & 30  \\
\hline\hline
   &  OA(\%)        &  94.24 &94.45 &{\bf 94.91} &94.53 &{94.13} &94.32 &94.11 &93.75 &94.05 &  94.51& 94.44 &  94.21\\
 PU  &  AA(\%)      &  93.58 &{\bf 94.36} &94.13 &93.57 &{92.29} &93.61 &93.02 &92.58 &93.76  & 93.54 & 93.49 & 93.34 \\
   &  $\kappa$(\%)  &  92.21 &92.53 &{\bf 93.15} &92.63 &{92.06} &92.35 &92.05 &91.56 &92.01   & 92.58 & 92.49 & 92.19\\
\hline
   &  OA(\%)       & 88.87 & {89.40}  &89.10 &88.89 &88.94 &88.72 &88.95 &89.66  & {\bf 90.13} & 90.06 & 88.85& 87.30 \\
  IP &  AA(\%)     & 93.63 & {94.11}  &93.22 &94.03 &93.77 &93.80 & 93.41& 94.25  & 94.10 & {\bf 94.65} & 94.15& 92.65 \\
   &  $\kappa$(\%) & 87.28 & {87.85}  &87.54 &87.28 &87.33 &87.13 &87.33 &88.16  & {\bf 88.67} & 88.62 & 87.23& 85.47 \\
\hline
   &  OA(\%)       &  89.57  & {\bf 89.74} &86.93 &{87.29} & 88.73 & 88.04 &88.66 &88.99 &87.73   & 88.75 & 89.09 & 86.60 \\
  HS &  AA(\%)     &  91.30  & {\bf 91.57}&88.94 & {89.47}& 90.50& 90.13& 90.33& 90.53  & 89.52 & 90.37 & 90.91& 88.72\\
   &  $\kappa$(\%) &  88.68  & {\bf 88.87}&85.81 & {86.20}& 87.76 & 87.01& 87.69& 88.04&  86.68  & 87.79 & 88.15 & 85.45\\
\bottomrule[1pt]
\end{tabular}
\end{center}
\vspace{-0.5cm}
\end{table*}

\subsection{Comparative Analysis of Models Trained with Different Backbone CNNs}

The choice of backbone CNN emerges as a pivotal factor influencing the quality of extracted environment-related and category-related features, thereby significantly impacting the classification performance of hyperspectral images. In a series of experiments, we systematically evaluate the performance of the proposed method with distinct backbone CNNs, namely, 3-D CNN, PResNet, and HybridSN, with their respective structures set as subsection \ref{subsubsec:baseline} shows.

Tables \ref{table:pavia_cnns}, \ref{table:indian_cnns}, and \ref{table:houston2013_cnns} present a comparative analysis of the proposed method against Vanilla CNNs across three datasets. Notable observations from these results include:

First, {\bf Backbone Influence on Performance}. The performance achieved with PResNet and HybridSN as backbones surpasses that with 3-D CNN. For instance, on the Pavia University dataset, the proposed method achieves 94.30\% and 94.45\% accuracy with PResNet and HybridSN, outperforming the 88.62\% accuracy obtained with 3-D CNN. Similar trends are observed for Indian Pines and Houston2013 datasets.
Then, {\bf Enhancement through HyperDID}. The proposed HyperDID significantly improves the performance of Vanilla CNNs. Across all datasets, the method achieves notable improvements: 0.57\% to 8.82\% with 3-D CNN, 4.19\% to 5.85\% with PResNet, and 2.85\% to 10.68\% with HybridSN. This underscores the effectiveness of the proposed approach in elevating the classification accuracy of hyperspectral images.

These findings underscore the importance of selecting an appropriate backbone CNN architecture and highlight the substantial performance gains achievable through the proposed HyperDID, particularly when integrated with advanced backbone architectures like PResNet and HybridSN.

\begin{table}[t]
\begin{center}
\caption{Classification accuracies (OA, AA, and $\kappa$) of the proposed method with different backbone CNNs over Pavia University data.}
\label{table:pavia_cnns}
%\vspace{-1ex}
\begin{tabular}{ c | c || c c c}
\toprule[1pt]
\multirow{2}{*}{Data}     &  \multirow{2}{*}{Metrics} &  \multicolumn{3}{c}{CNN Backbone}  \\
\cline{3-5}
     &   &  {3-D CNN}&  {PResNet}&  {HybridSN}   \\
\hline\hline
   &  OA(\%)        &  87.52 & 90.11&90.27   \\
 Vanilla  &  AA(\%)      &  89.01 & 89.43&91.79    \\
   &  $\kappa$(\%)  &  83.37 & 86.68&87.03    \\
\hline
   &  OA(\%)        &  88.62 & 94.30&94.45   \\
 Proposed  &  AA(\%)      &  90.76 & 94.02&94.36    \\
   &  $\kappa$(\%)  &  85.06 & 92.37&92.53    \\
\bottomrule[1pt]
\end{tabular}
\end{center}
\vspace{-0.5cm}
\end{table}

\begin{table}[t]
\begin{center}
\caption{Classification accuracies (OA, AA, and $\kappa$) of the proposed method with different backbone CNNs over Indian Pines data.}
\label{table:indian_cnns}
%\vspace{-1ex}
\begin{tabular}{ c | c || c c c}
\toprule[1pt]
\multirow{2}{*}{Data}     &  \multirow{2}{*}{Metrics} &  \multicolumn{3}{c}{CNN Backbone}  \\
\cline{3-5}
     &   &  {3-D CNN}&  {PResNet}&  {HybridSN}   \\
\hline\hline
   &  OA(\%)        &  77.22 & 82.97&78.72   \\
 Vanilla  &  AA(\%)      &  86.83 & 90.19&88.15    \\
   &  $\kappa$(\%)  &  74.21 & 80.65&75.81    \\
\hline
   &  OA(\%)        &  {86.04} &88.82 &89.40    \\
  Proposed &  AA(\%)      &  90.80 &93.49 &{94.11}    \\
   &  $\kappa$(\%)  &  {84.08} &87.23 &87.85    \\
\bottomrule[1pt]
\end{tabular}
\end{center}
\vspace{-0.5cm}
\end{table}

\begin{table}[t]
\begin{center}
\caption{Classification accuracies (OA, AA, and $\kappa$) of the proposed method with different backbone CNNs over Houston2013 data.}
\label{table:houston2013_cnns}
%\vspace{-1ex}
\begin{tabular}{ c | c || c c c}
\toprule[1pt]
\multirow{2}{*}{Data}     &  \multirow{2}{*}{Metrics} &  \multicolumn{3}{c}{CNN Backbone}  \\
\cline{3-5}
     &   &  {3-D CNN}&  {PResNet}&  {HybridSN}   \\
\hline\hline
   &  OA(\%)        &  {84.71} &85.59 &86.89    \\
  Vanilla &  AA(\%)      &  85.53 &87.45 &{88.92}    \\
   &  $\kappa$(\%)  &  {83.40} &84.35 &85.77    \\
\hline
   &  OA(\%)       & 85.28  &90.11 &89.74    \\
  Proposed &  AA(\%)     & 86.41  &90.83 &91.57    \\
   &  $\kappa$(\%) & 84.02  &89.26 &88.87    \\
\bottomrule[1pt]
\end{tabular}
\end{center}
\vspace{-0.6cm}
\end{table}

\subsection{Comparative Analysis of Models Trained with Different Spatial Neighbor Sizes}

\begin{figure*}[t]
\centering
 \subfigure[]{\label{subfig:pavia}\includegraphics[width=0.32\linewidth]{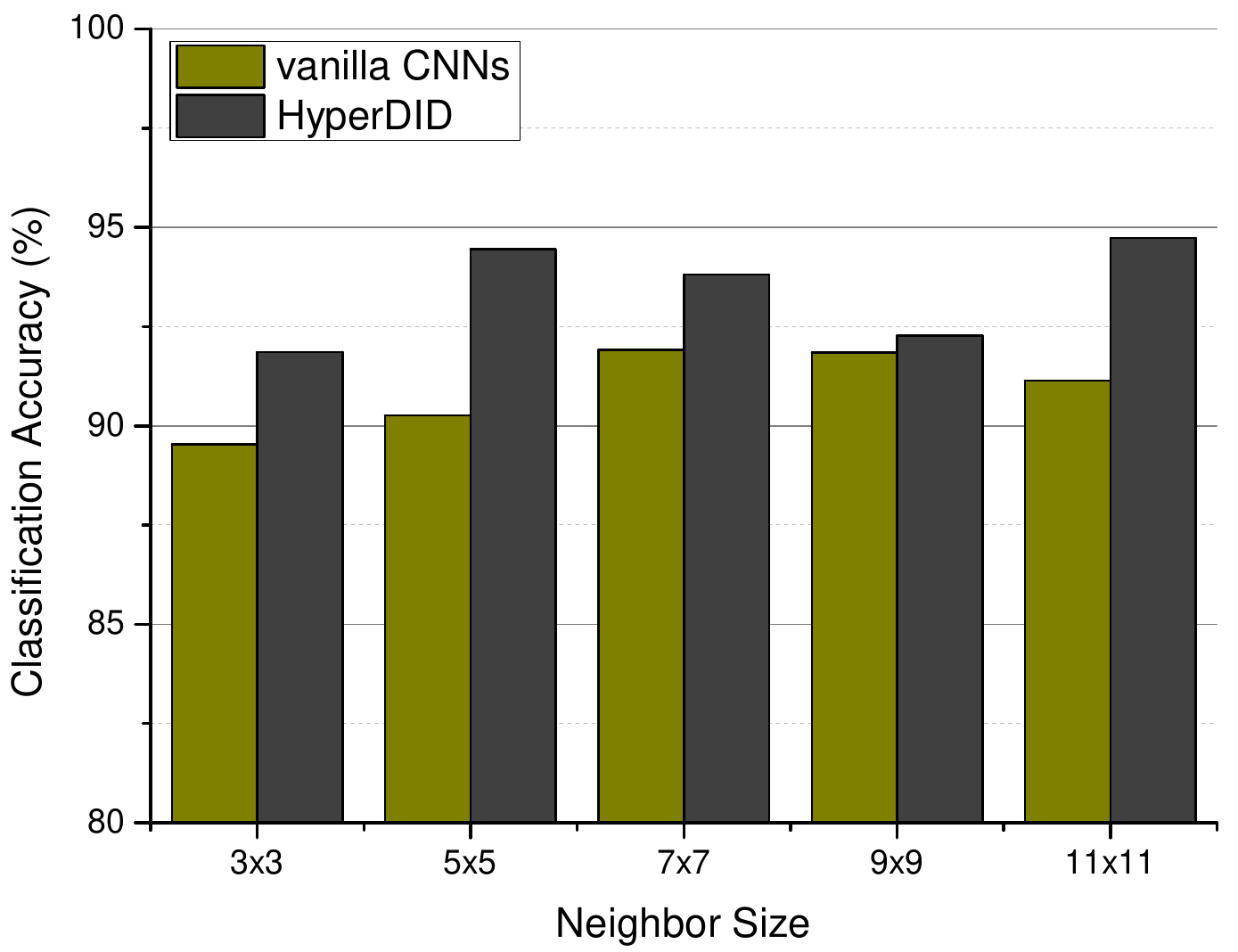}}
 \subfigure[]{\label{subfig:pavia_label}\includegraphics[width=0.32\linewidth]{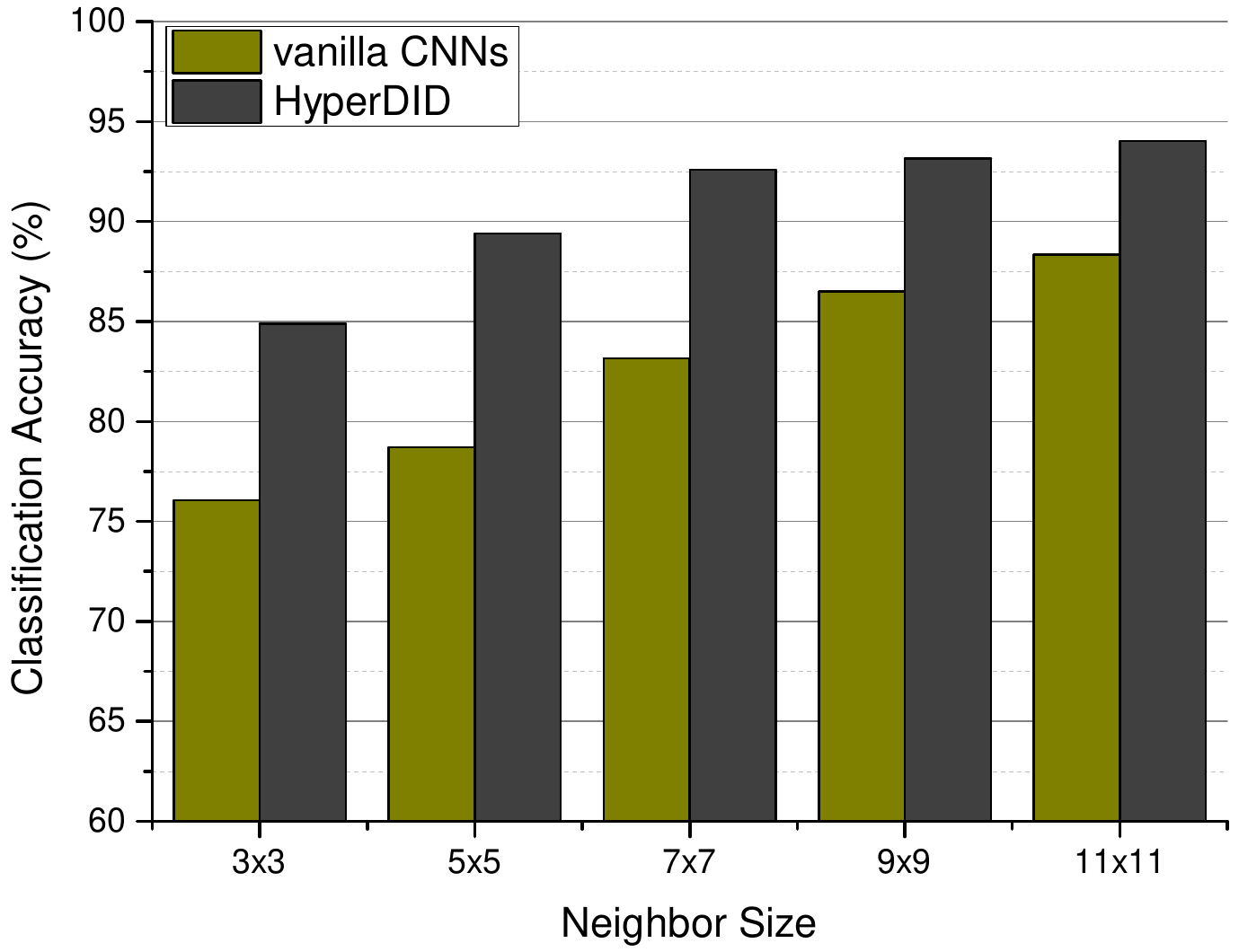}}
 \subfigure[]{\label{subfig:mapping}\includegraphics[width=0.32\linewidth]{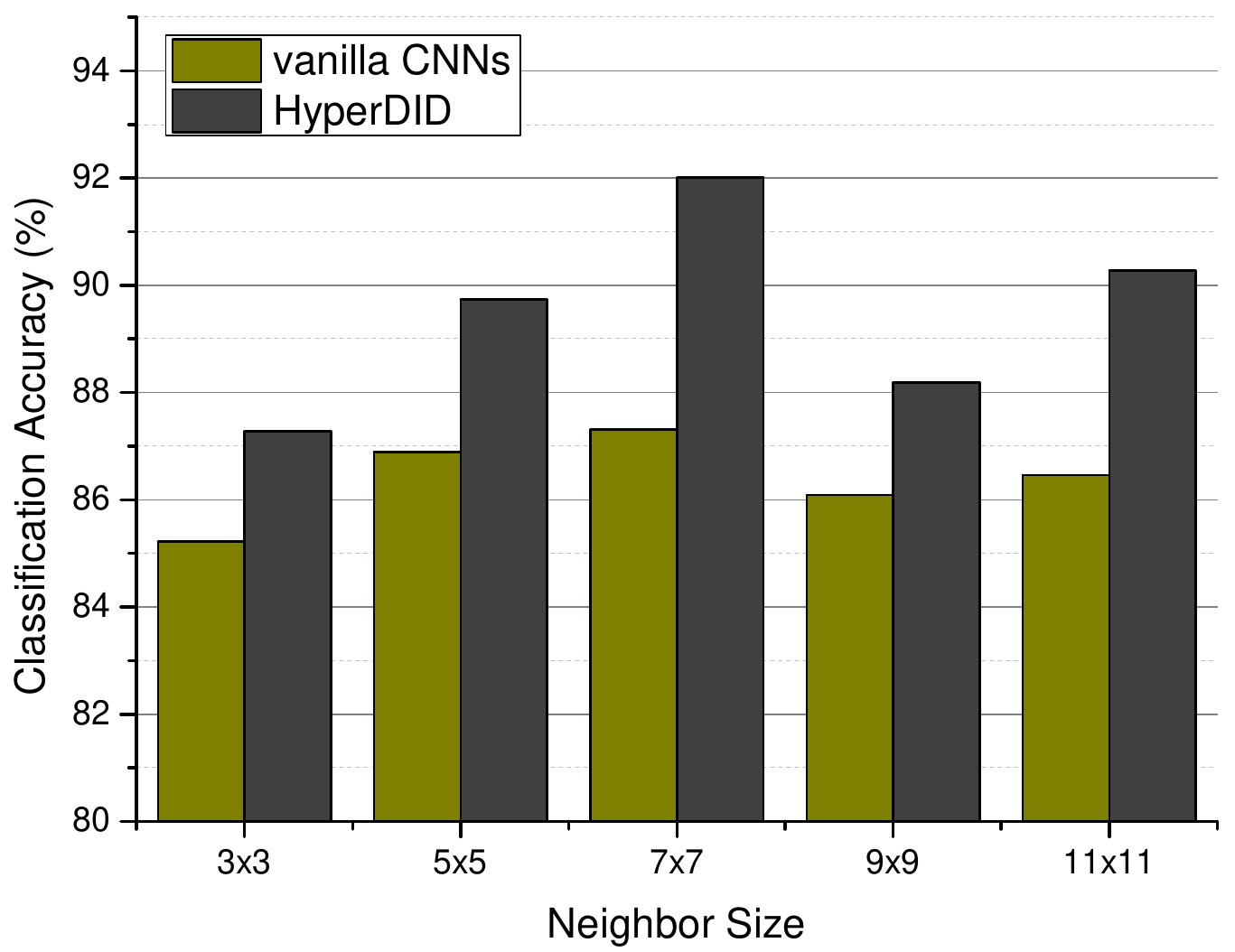}}
   \caption{Classification performance with different spatial neighbor sizes over (a) Pavia University; (b) Indian Pines; (c) Houston2013.}
\label{fig:number}
\vspace{-0.5cm}
\end{figure*}

The significance of spatial neighbors in enhancing the accuracy of hyperspectral image classification cannot be overstated, as they contribute a wealth of spatial information crucial for the task. The number of neighbor sizes emerges as a key factor influencing classification performance, a notion explored through additional comparison experiments in this study. Spatial neighbor sizes ranging from $3\times 3$ to $11\times 11$ were selected for evaluation, and Fig. \ref{fig:number} illustrates the evolving trends in classification accuracies across three datasets.

A consistent observation is that training with larger spatial neighbor sizes tends to yield improved performance. However, it is essential to strike a balance, as excessively large neighbor sizes can introduce irrelevant information, negatively impacting the training process and hindering the extraction of discriminative features. The figure illustrates that, for Indian Pines and Pavia University datasets, HyperDID achieves optimal performance with spatial neighbors of $11\times 11$, yielding high accuracy metrics. Conversely, for Houston 2013 data, the best performance is attained with a neighbor size of $7\times 7$. These findings underscore the importance of carefully selecting spatial neighbor sizes to achieve an optimal trade-off between capturing relevant spatial information and avoiding information redundancy during hyperspectral image classification.

\subsection{Comparative Analysis of Models Trained with Different Number of Training Samples}

The earlier sections of our study primarily delved into experiments conducted with predefined sets of training and testing samples, as outlined in Tables \ref{table:pavia}, \ref{table:indian}, and \ref{table:houston2013}. This subsection extends the evaluation of the HyperDID method by considering varying numbers of training samples.

\begin{figure*}[t]
\centering
 \subfigure[]{\label{subfig:pavia}\includegraphics[width=0.32\linewidth]{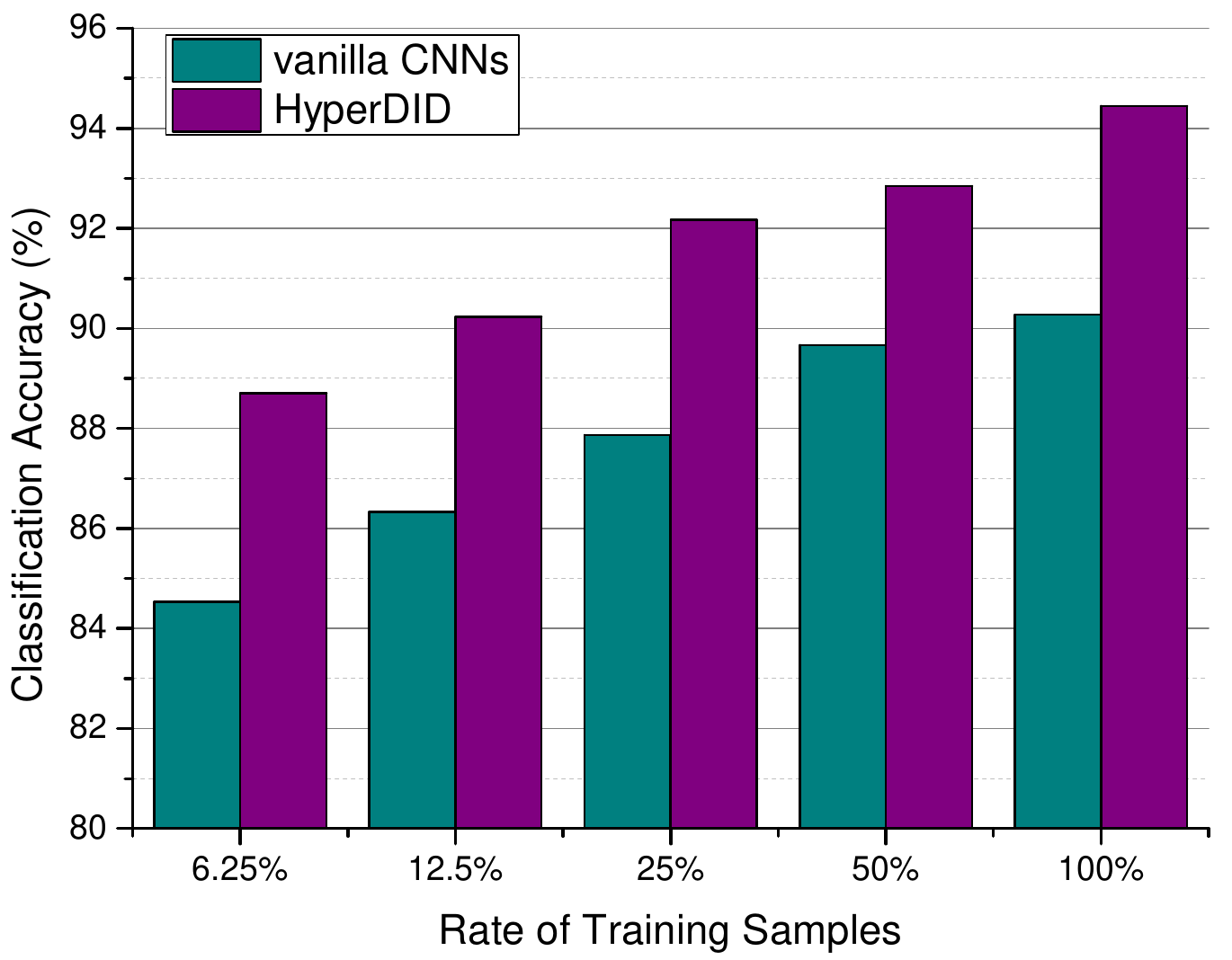}}
 \subfigure[]{\label{subfig:pavia_label}\includegraphics[width=0.32\linewidth]{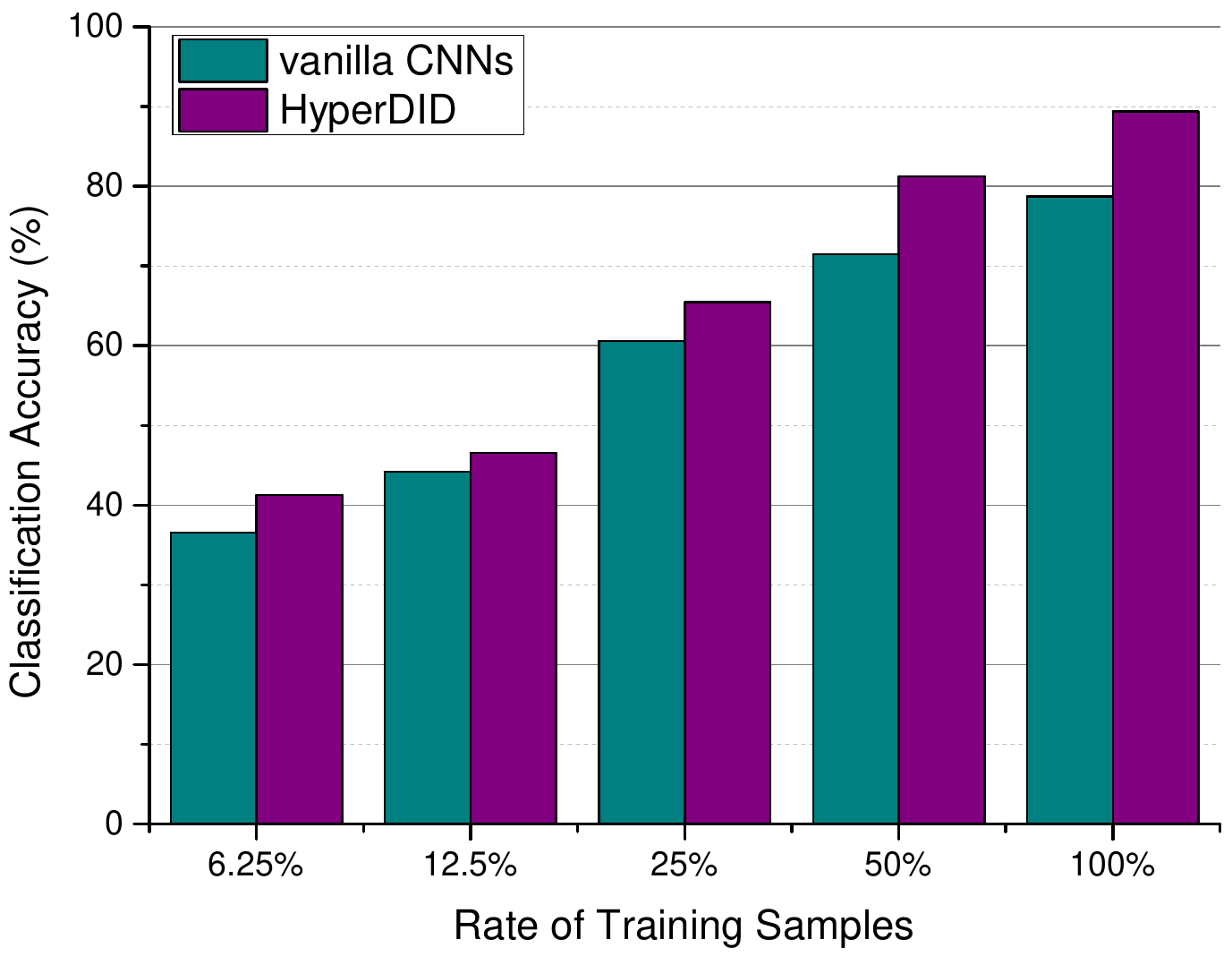}}
 \subfigure[]{\label{subfig:mapping}\includegraphics[width=0.32\linewidth]{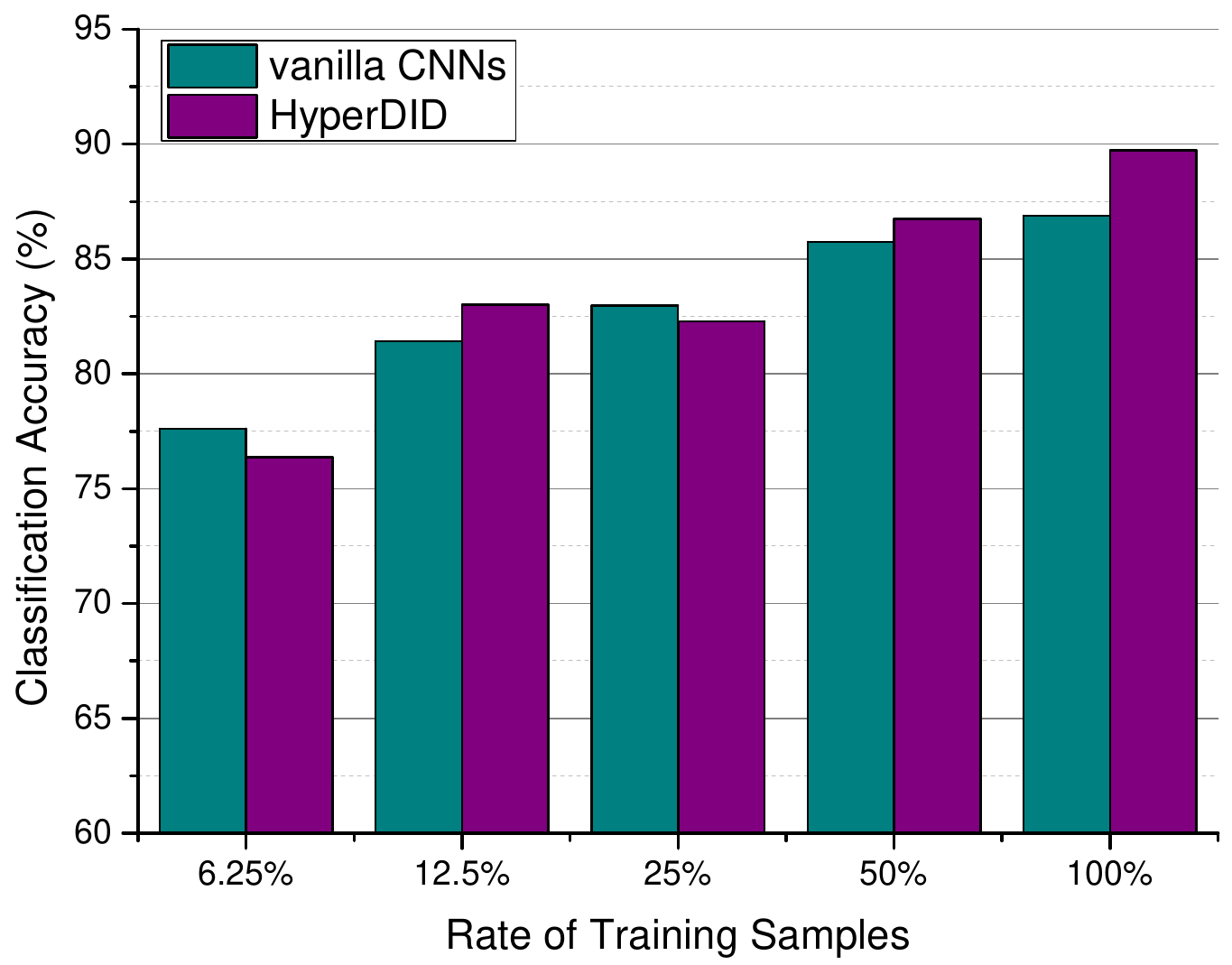}}
   \caption{Classification performance with different rate of training samples over (a) Pavia University; (b) Indian Pines; (c) Houston2013.}
\label{fig:rate}
\vspace{-0.5cm}
\end{figure*}

In the initial experiments, we utilized 3921 training and 40002 testing samples for Pavia University data, 695 training and 9671 testing samples for Indian Pines data, and 2832 training and 12197 testing samples for Houston2013 data, as indicated in Tables I-III. In this subsequent set of experiments, we systematically selected 6.25\%, 12.5\%, 25\%, 50\%, and 100\% of the original training samples across these datasets to assess performance under different training sample scenarios. For example, over Pavia University data, the selected training samples ranged from 245 to 3921, with similar sample selections applied to Indian Pines and Houston2013 data. The trends in classification performance corresponding to these varied numbers of training samples are visually depicted in Fig. \ref{fig:rate}.

Noteworthy results highlight a substantial improvement in accuracies achieved by the proposed HyperDID method compared to vanilla CNNs. Pavia University data saw an accuracy enhancement of approximately 3\%-4\%. While Houston2013 data observed an improvement of about 1.5\%-3\%. It should be noted that the performance decreases when the rate is set to 6.25\% and 25\%. The reason may be that unstable training process with less training samples. Particularly remarkable is the performance on Indian Pines data, where the accuracy experienced an increase of more than 10\%. This enhancement is attributed to HyperDID's ability to decompose environment-related and category-related features, thereby improving discrimination of category-related features and mitigating the impact of environmental factors on hyperspectral image classification.

Moreover, the classification performance of the learned model showed a significant enhancement with an increase in the number of training samples. This escalating trend indicates that more training samples provide additional information for the deep model to learn, enabling it to extract more discriminative features for improved hyperspectral image classification. This observation underscores the scalability and learning capacity of HyperDID, emphasizing the critical role of ample training data in achieving superior classification performance in hyperspectral image analysis.

\subsection{Comparisons with Other State-of-the-Art Methods}

The quantitative classification results presented in Tables \ref{table:pavia_comparison}, \ref{table:indian_comparison}, and \ref{table:houston2013_comparison} for the Indian Pines, Pavia University, and Houston2013 datasets, respectively, underscore the superior performance of the proposed HyperDID under consistent experimental setups. 

Table \ref{table:pavia_comparison} showcases the exceptional performance of HyperDID over Pavia University data with an accuracy of 94.45\%OA, 94.36\%AA, and 92.53\% kappa, outperforming state-of-the-art methods such as traditional machine learning techniques like SVM (83.76\%OA), CNNs like 3-D CNNs (87.52\%OA), PResNet (90.11\%OA), HybridSN (90.27\%OA), RNNs (80.61\%), GCNs (miniGCN with 83.23\%OA), and Transformers like ViT (86.27\%), SpectralFormer (90.04\%), and SSFTTNet (82.56\%).
For the Indian Pines dataset, HyperDID achieves an accuracy of 89.40\%OA, 94.11\%AA, and 87.85\% kappa, surpassing alternative methods including SVM (76.53\%OA), 3-D CNNs (77.22\%OA), PResNet (82.97\%OA), HybridSN (78.72\%OA), RNN (81.11\%), miniGCN (74.71\%OA), ViT (65.16\%), SpectralFormer (83.38\%), and SSFTTNet (80.29\%). Similarly, over the Houston2013 dataset, HyperDID outperforms its counterparts.

\begin{figure*}[t]
\centering
 \subfigure[]{\label{subfig:pavia_1}\includegraphics[width=0.16\linewidth]{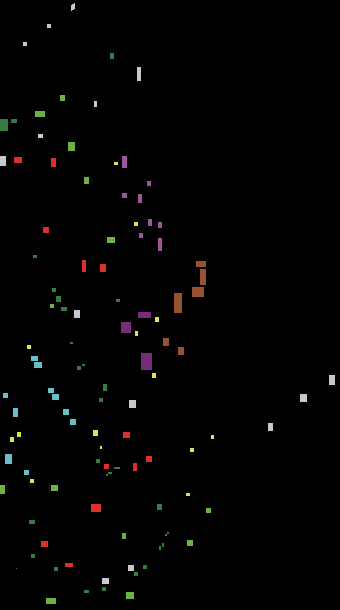}}
 \subfigure[]{\label{subfig:pavia_2}\includegraphics[width=0.16\linewidth]{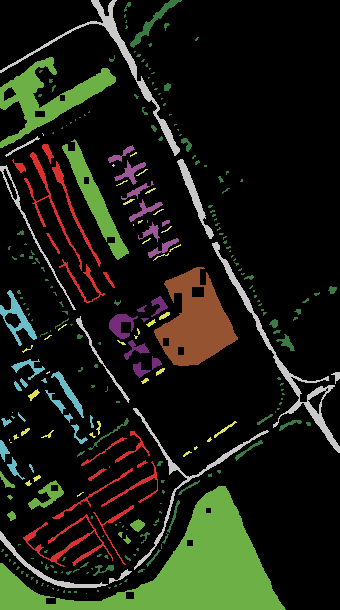}}
 \subfigure[]{\label{subfig:pavia_3}\includegraphics[width=0.16\linewidth]{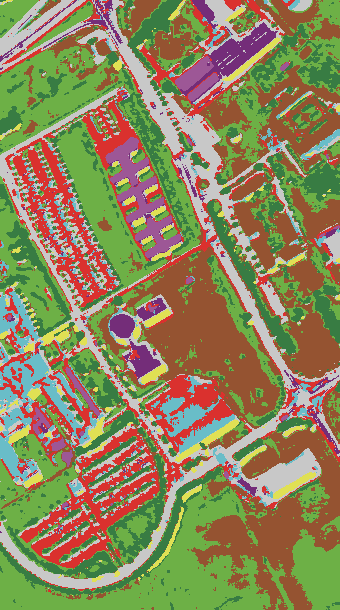}}
\subfigure[]{\label{subfig:pavia_4}\includegraphics[width=0.16\linewidth]{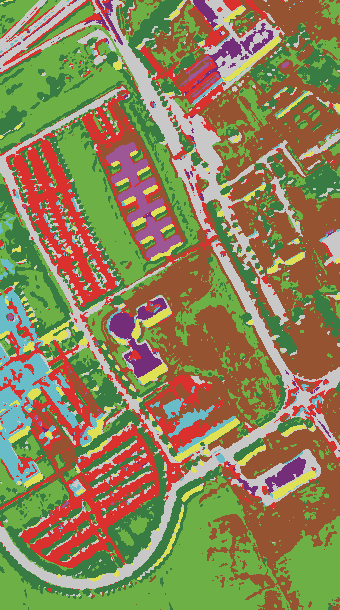}}
\subfigure[]{\label{subfig:pavia_5}\includegraphics[width=0.16\linewidth]{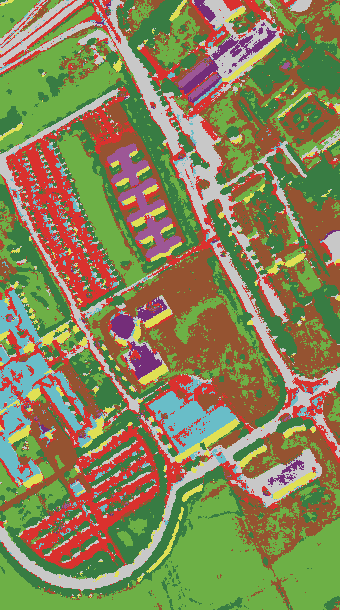}}
\subfigure[]{\label{subfig:pavia_6}\includegraphics[width=0.16\linewidth]{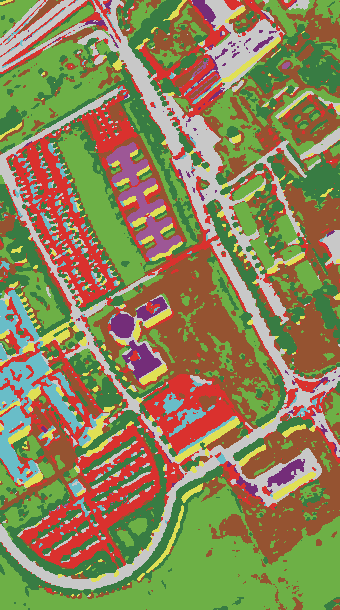}}
\subfigure[]{\label{subfig:pavia_7}\includegraphics[width=0.16\linewidth]{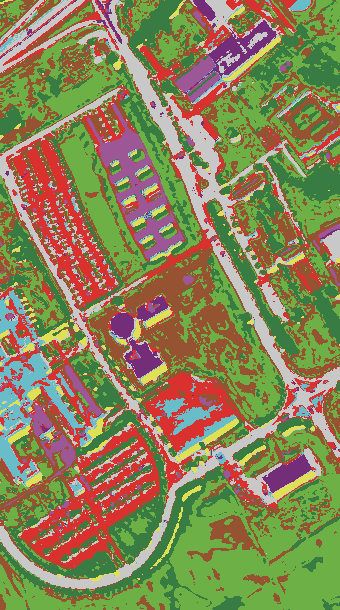}}
\subfigure[]{\label{subfig:pavia_8}\includegraphics[width=0.16\linewidth]{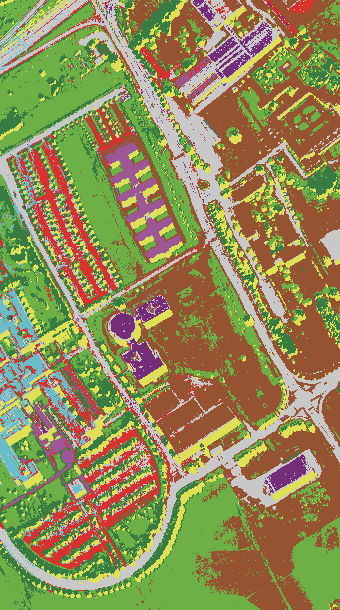}}
\subfigure[]{\label{subfig:pavia_9}\includegraphics[width=0.16\linewidth]{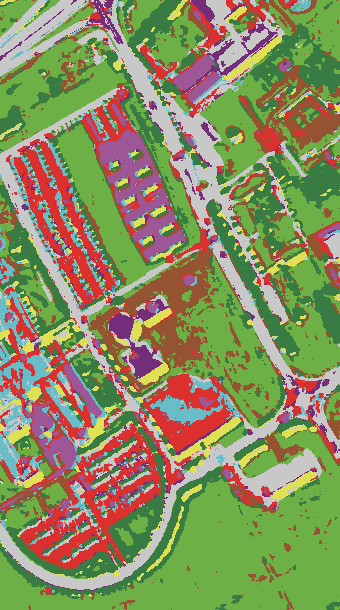}}
\subfigure[]{\label{subfig:pavia_10}\includegraphics[width=0.16\linewidth]{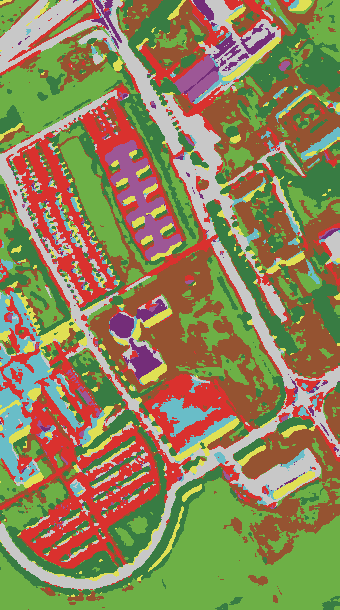}}
\subfigure[]{\label{subfig:pavia_11}\includegraphics[width=0.16\linewidth]{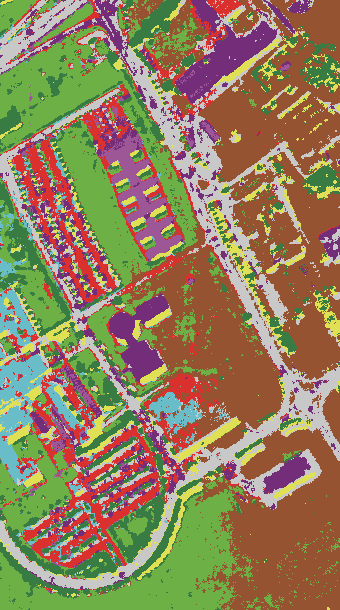}}
\subfigure[]{\label{subfig:pavia_12}\includegraphics[width=0.16\linewidth]{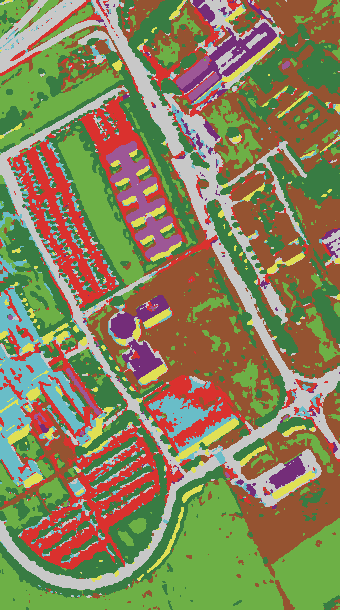}}
   \caption{Pavia University data. (a) Training; (b) Testing; (c) SVM; (d) 3-D CNN; (e) PResNet; (f) HybridSN; (g) RNN; (h) miniGCN; (i) ViT; (j) SpectralFormer; (k) SSFTTNet; (l) HyperDID.}
\label{fig:pavia}
\vspace{-0.5cm}
\end{figure*}

\begin{figure*}[t]
\centering
 \subfigure[]{\label{subfig:indian_1}\includegraphics[width=0.16\linewidth]{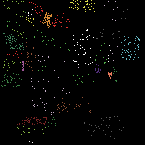}}
 \subfigure[]{\label{subfig:indian_2}\includegraphics[width=0.16\linewidth]{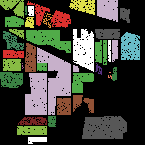}}
 \subfigure[]{\label{subfig:indian_3}\includegraphics[width=0.16\linewidth]{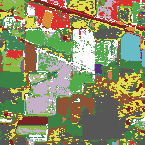}}
\subfigure[]{\label{subfig:indian_4}\includegraphics[width=0.16\linewidth]{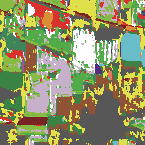}}
\subfigure[]{\label{subfig:indian_5}\includegraphics[width=0.16\linewidth]{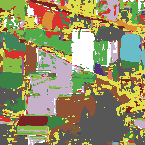}}
\subfigure[]{\label{subfig:indian_6}\includegraphics[width=0.16\linewidth]{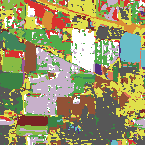}}
\subfigure[]{\label{subfig:indian_7}\includegraphics[width=0.16\linewidth]{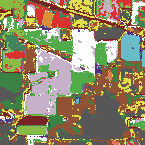}}
\subfigure[]{\label{subfig:indian_8}\includegraphics[width=0.16\linewidth]{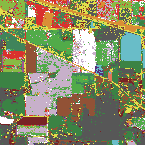}}
\subfigure[]{\label{subfig:indian_9}\includegraphics[width=0.16\linewidth]{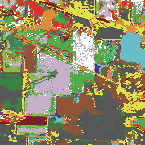}}
\subfigure[]{\label{subfig:indian_10}\includegraphics[width=0.16\linewidth]{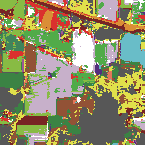}}
\subfigure[]{\label{subfig:indian_11}\includegraphics[width=0.16\linewidth]{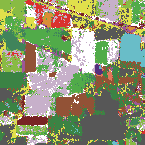}}
\subfigure[]{\label{subfig:indian_12}\includegraphics[width=0.16\linewidth]{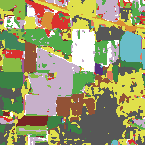}}

   \caption{Indian Pines data. (a) Training; (b) Testing; (c) SVM; (d) 3-D CNN; (e) PResNet; (f) HybridSN; (g) RNN; (h) miniGCN; (i) ViT; (j) SpectralFormer; (k) SSFTTNet; (l) HyperDID.}
\label{fig:indian}
\vspace{-0.5cm}
\end{figure*}

\begin{figure*}[t]
\centering
 \subfigure[]{\label{subfig:houston2013_1}\includegraphics[width=0.49\linewidth]{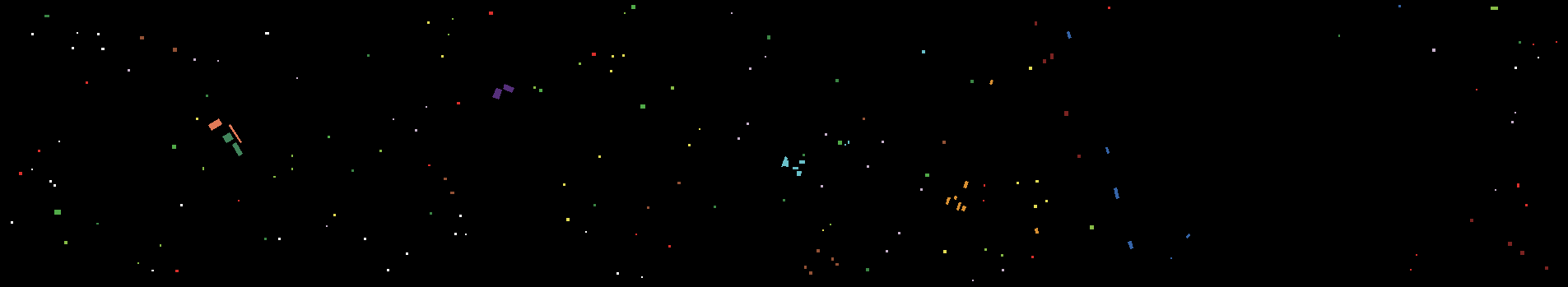}}
 \subfigure[]{\label{subfig:houston2013_2}\includegraphics[width=0.49\linewidth]{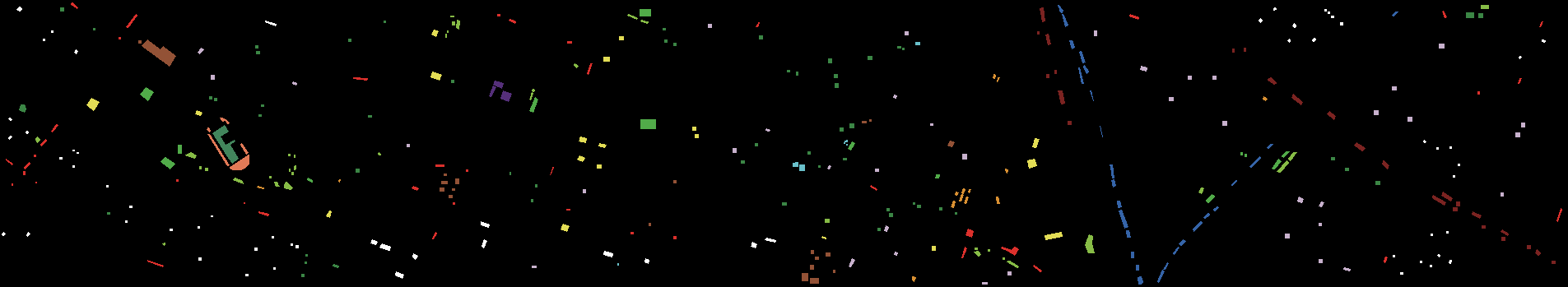}}
 \subfigure[]{\label{subfig:houston2013_3}\includegraphics[width=0.49\linewidth]{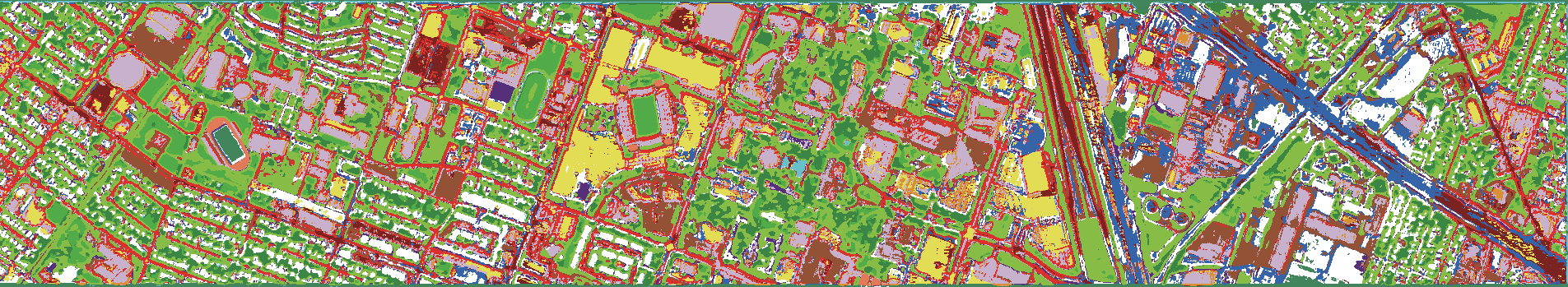}}
\subfigure[]{\label{subfig:houston2013_4}\includegraphics[width=0.49\linewidth]{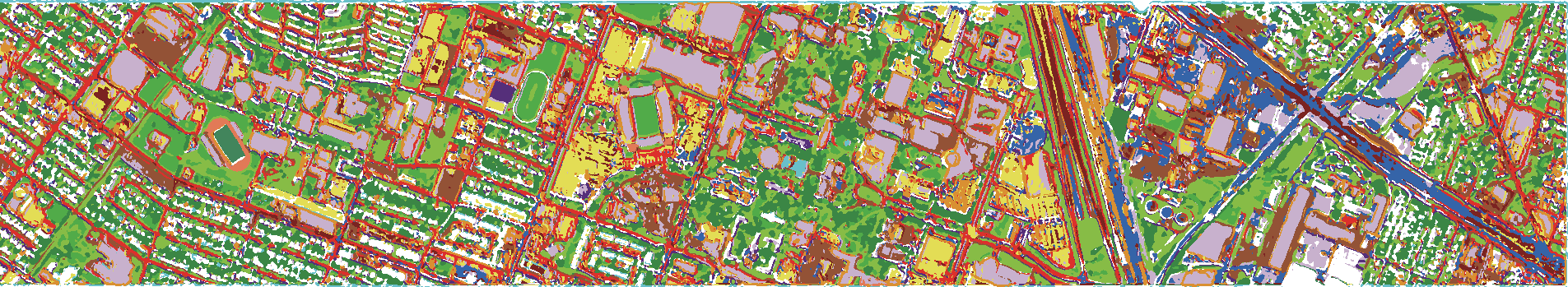}}
\subfigure[]{\label{subfig:houston2013_5}\includegraphics[width=0.49\linewidth]{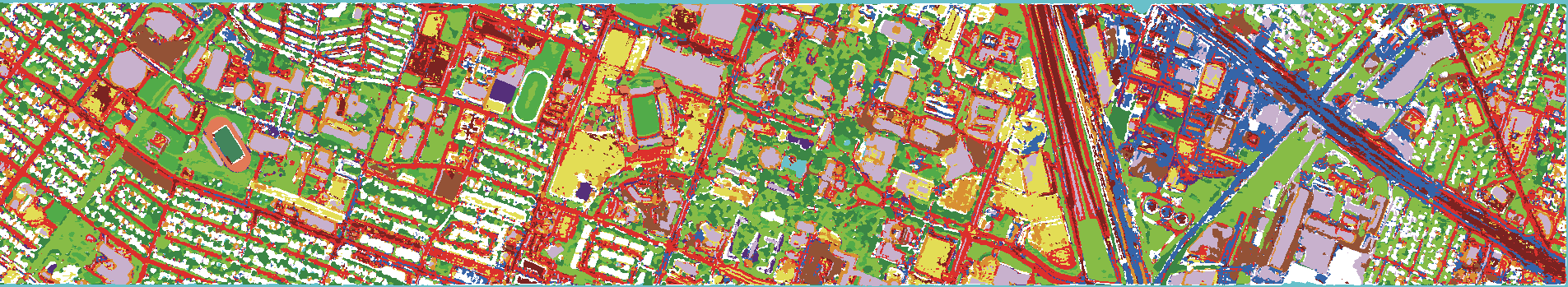}}
\subfigure[]{\label{subfig:houston2013_6}\includegraphics[width=0.49\linewidth]{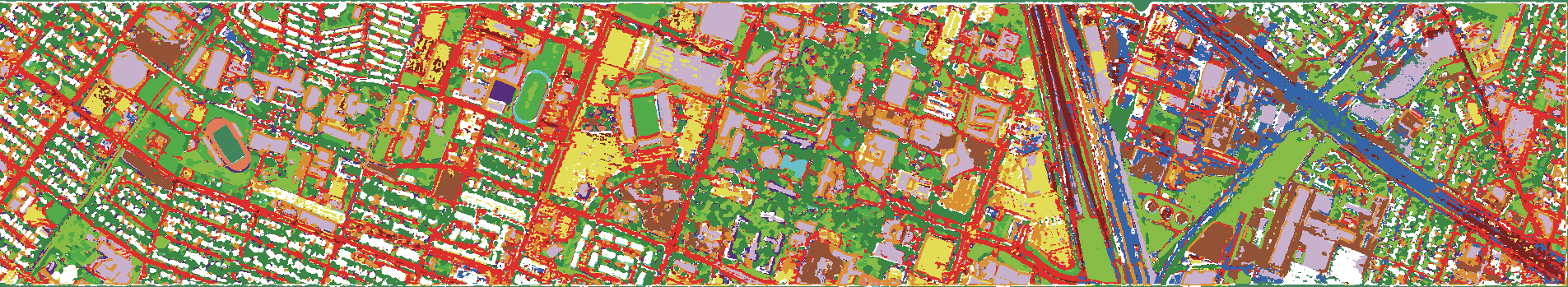}}
\subfigure[]{\label{subfig:houston2013_7}\includegraphics[width=0.49\linewidth]{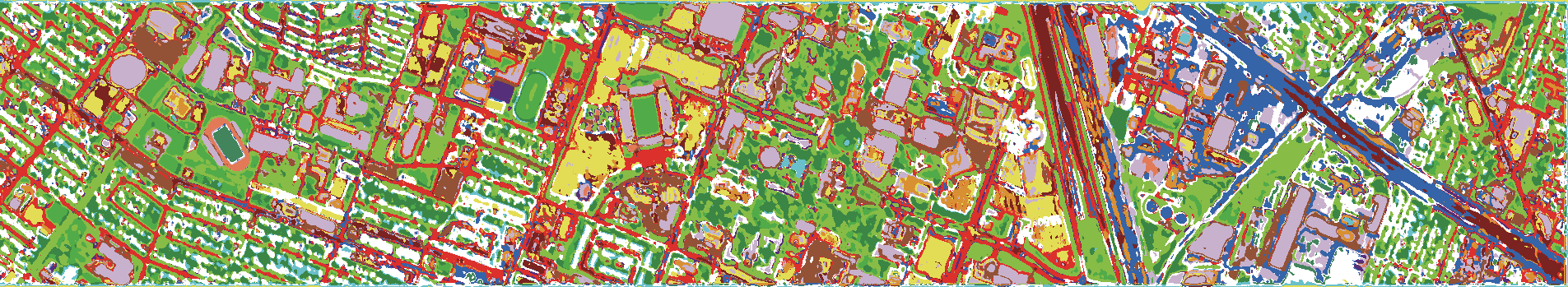}}
\subfigure[]{\label{subfig:houston2013_8}\includegraphics[width=0.49\linewidth]{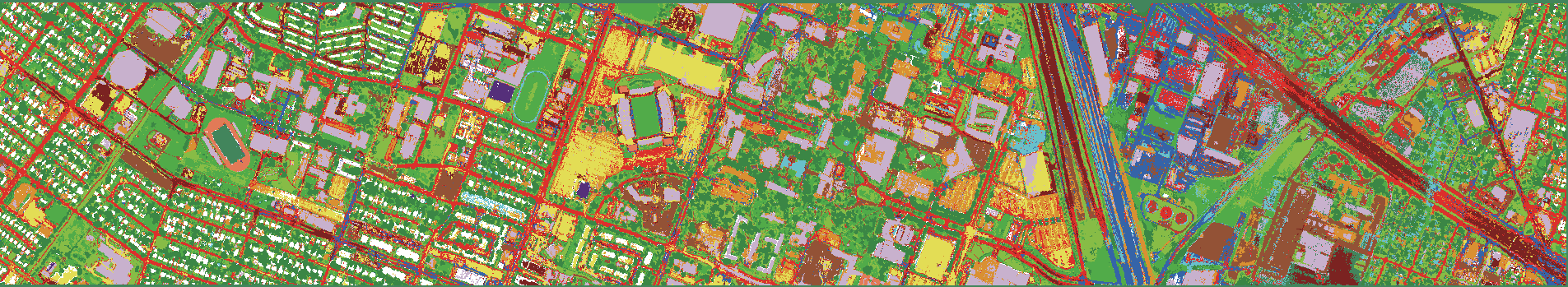}}
\subfigure[]{\label{subfig:houston2013_9}\includegraphics[width=0.49\linewidth]{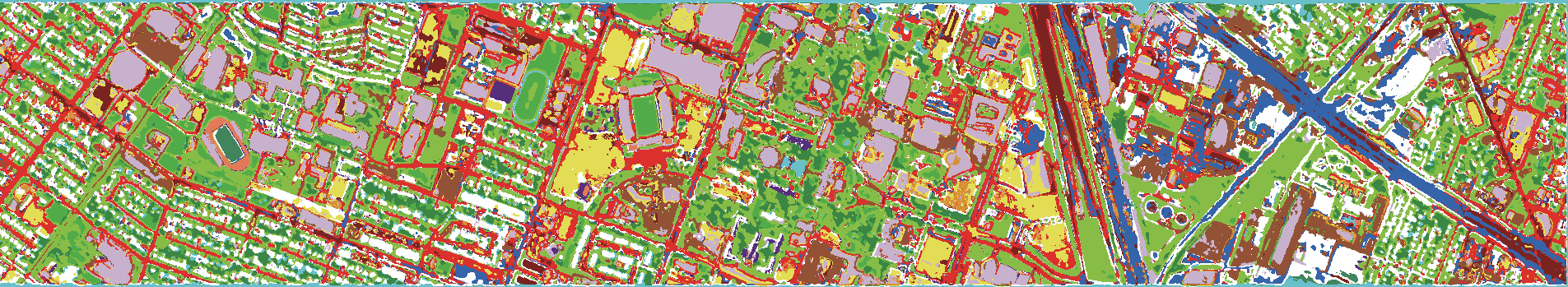}}
\subfigure[]{\label{subfig:houston2013_10}\includegraphics[width=0.49\linewidth]{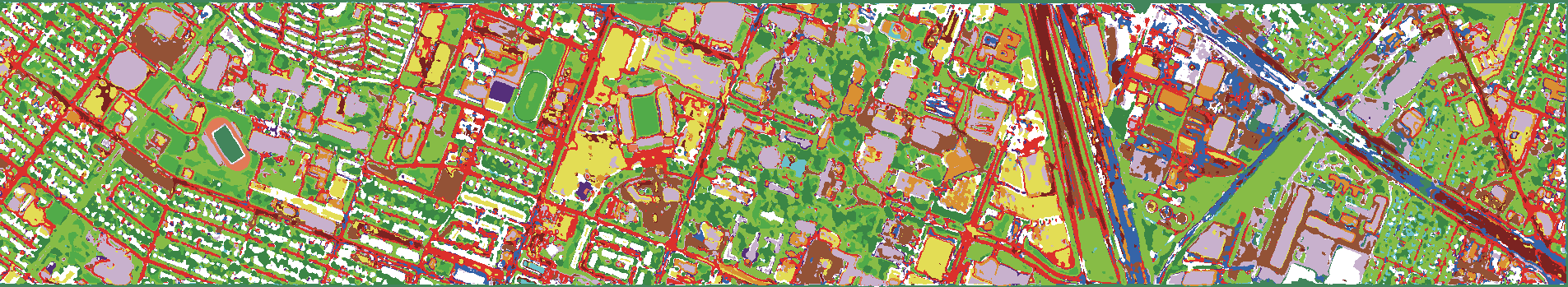}}
\subfigure[]{\label{subfig:houston2013_11}\includegraphics[width=0.49\linewidth]{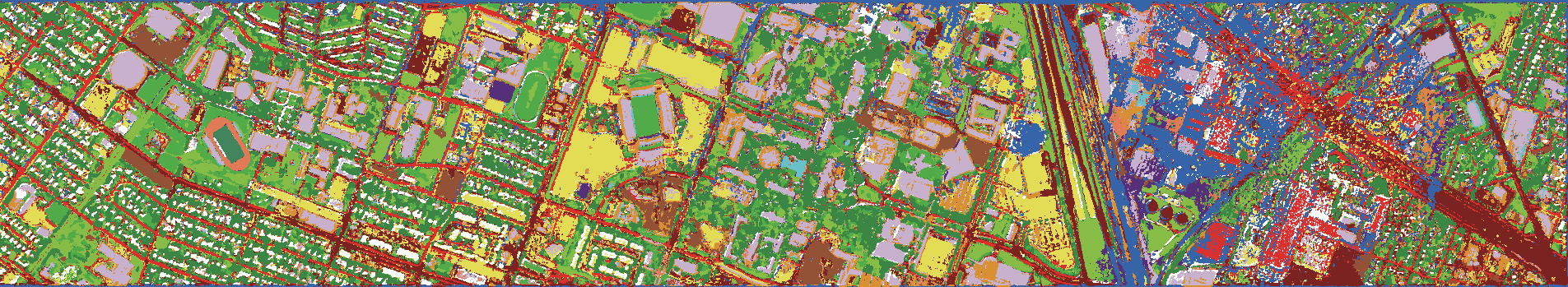}}
\subfigure[]{\label{subfig:houston2013_12}\includegraphics[width=0.49\linewidth]{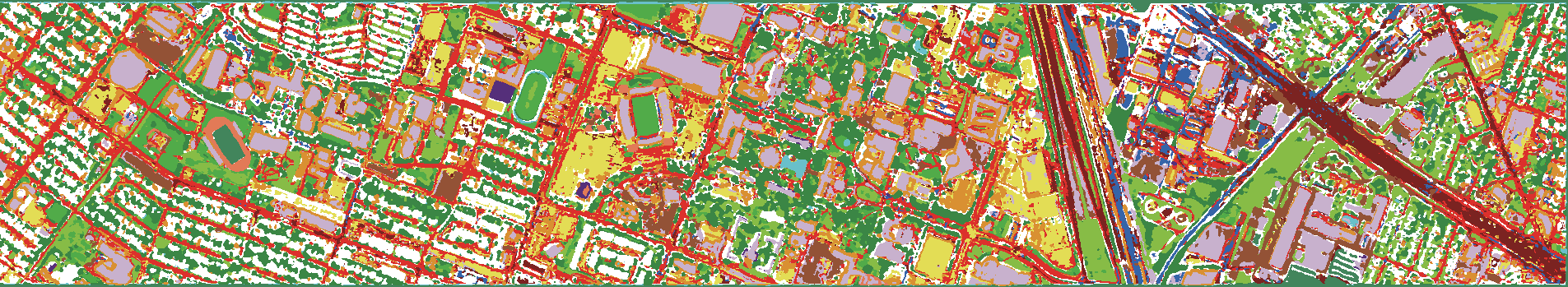}}

   \caption{Houston2013 data. (a) Training; (b) Testing; (c) SVM; (d) 3-D CNN; (e) PResNet; (f) HybridSN; (g) RNN; (h) miniGCN; (i) ViT; (j) SpectralFormer; (k) SSFTTNet; (l) HyperDID.}
\label{fig:houston2013}
\vspace{-0.5cm}
\end{figure*}

A qualitative evaluation through visualization of classification maps in Figs. \ref{fig:pavia}, \ref{fig:indian}, and \ref{fig:houston2013} affirms the effectiveness of HyperDID in mitigating salt-and-pepper noise-induced classification errors. This leads to a substantial reduction in overall inaccuracies and a notable improvement in classification accuracy. The method's ability to suppress the impact of challenging noise patterns underscores its robustness, resulting in more reliable and accurate classification outcomes.

In summary, HyperDID, leveraging intrinsic information from hyperspectral images, demonstrates a significant performance boost compared to state-of-the-art methods. Its ability to achieve superior accuracy across diverse datasets positions it as a robust and effective solution for hyperspectral image classification tasks.

\begin{table*}[t]
\setlength{\aboverulesep}{-0.1pt}
\setlength{\belowrulesep}{-0.1pt}
\begin{center}
\caption{Classification accuracies (OA, AA, and $\kappa$) of different methods achieved on the Pavia University data.}
\label{table:pavia_comparison}
%\vspace{-1ex}
\begin{tabular}{ c || c | c c c | c | c | c c c || c}
\toprule[1pt]
\multirow{2}{*}{Methods}     &  \multirow{2}{*}{SVM} &  \multicolumn{3}{c|}{CNNs} &  \multirow{2}{*}{RNN} & \multirow{2}{*}{miniGCN} & \multicolumn{3}{c||}{Transformers} &\multirow{2}{*}{HyperDID} \\
 \cmidrule(lr){3-5} \cmidrule(lr){8-10}
    &   &  { 3-D CNN}&  {PResNet }&  { HybridSN} &   &  & { ViT} & {SpectralFormer} & SSFTTNet & \\
\hline\hline
 C1   &  77.08  & 82.95  &80.11 &83.49 &85.56 & 91.55 &82.55 &84.72 &75.89 & {\bf 93.51}\\
 C2   &  79.22  & 88.32  &94.81 &91.52 &75.00 & 84.62 &{\bf 96.57} &95.86 &81.46 & 95.97\\
 C3   &  77.52  & 73.55  &{85.62} &81.76 &71.63 & 74.27 &56.42 &66.72 &75.54 & {\bf 88.76}\\
 C4   &  94.61  & 93.96  &98.49 &{98.80} &94.16 & 71.22 &95.98 &96.53 &83.72 & {\bf 99.04}\\
 C5   &  98.74  & 99.55  &99.91 &100.0 &91.46 & 99.55 &93.62 &99.19 &{\bf 100.0} & 99.37\\
 C6   &  {\bf 93.68}  & 79.11  &79.48 &85.39 &72.66 & 67.61 &49.43 &73.16 &83.44 & 87.93\\
 C7   &  85.12  & 90.21  &79.00 &91.85 &92.25 & 86.75 &79.61 &79.71 &{\bf 99.80} & 93.37\\
 C8   &  93.82  & {\bf 98.63}  &91.77 &94.41 &95.54 & 86.15 &93.79 &97.74 &88.61 & {94.02}\\
 C9   &  92.58  & 94.84  &95.72 &98.87 &93.21 & {\bf 100.0} &91.07 &93.33 &95.60 & 97.23\\
                                                                                                             
 \hline\hline
 {OA(\%)}      & 83.76  & 87.52 & 90.11 &90.27 &80.61 &83.23 &86.27 &90.04  & 82.56 & {\bf 94.45}\\
 
 {AA(\%)}      & 88.04  & 89.01 & 89.43 &91.79 &85.72 &84.64 &82.12 &87.44  & 87.12 & {\bf 94.36}\\
 
 {$\kappa$(\%)} & 78.98  & 83.37 & 86.68 &87.03 &74.83 &77.44 &81.16 &86.51  &77.30  & {\bf 92.53}\\
\bottomrule[1pt]
\end{tabular}
\end{center}
\vspace{-0.5cm}
\end{table*}

\begin{table*}[t]
\setlength{\aboverulesep}{-0.1pt}
\setlength{\belowrulesep}{-0.1pt}
\begin{center}
\caption{Classification accuracies (OA, AA, and $\kappa$) of different methods achieved on the Indian Pines data. }
\label{table:indian_comparison}
%\vspace{-1ex}
\begin{tabular}{ c || c | c c c | c | c | c c c || c}
\toprule[1pt]
\multirow{2}{*}{Methods}     &  \multirow{2}{*}{SVM} &  \multicolumn{3}{c|}{CNNs} &  \multirow{2}{*}{RNN} & \multirow{2}{*}{miniGCN} & \multicolumn{3}{c||}{Transformers} &\multirow{2}{*}{HyperDID} \\
 \cmidrule(lr){3-5} \cmidrule(lr){8-10}
    &   &  { 3-D CNN}&  {PResNet }&  { HybridSN} &   &  & { ViT} & {SpectralFormer} & SSFTTNet & \\
\hline\hline
 C1   & 72.11 &67.85     &72.76 & 66.04&72.76 &70.52  &59.68 &78.97 &{\bf 83.37} & {77.31} \\
 C2   & 71.43 &77.04     &87.50 & 79.46&84.82 &53.19  &37.76 &85.08 &71.39 &{\bf 89.92} \\
 C3   & 86.96 &93.48     &94.02 & {94.57}&78.26 &91.85  &55.43 &85.33 &94.57 &{\bf 97.83} \\
 C4   & 95.97 &92.84     &{ 94.85} & 91.28&88.14 &93.74  &65.10 &94.18 &93.93 &{\bf 96.87} \\
 C5   & 88.67 &83.21     &91.97 & 90.96&83.50 &95.12  &86.51 &84.36 &{95.38} &{\bf 97.56} \\
 C6   & 95.90 &98.63     &97.49 & {100.0}&91.80 &99.09  &97.95 &98.63 &98.40 &{\bf 100.0} \\
 C7   & 75.60 &74.51     &84.20 & 79.30&{87.58} &63.94  &51.85 &63.94 &72.66 &{\bf 89.98} \\
 C8   & 59.02 &62.66     &73.57 & 68.07&74.28 &68.40  &62.45 &{ 83.95} &71.63 &{\bf 87.34} \\
 C9   & 76.77 &69.68     &75.18 & 70.21&{75.35} &73.40  &41.13 &73.58 &50.18 &{\bf 79.79} \\
 C10   &99.38 &99.38     &100.0 & 100.0&99.38 &98.77  &96.30 &99.38 &98.15 &{\bf 100.0} \\
 C11   &93.33 &93.25      &93.41 & 76.94&93.89 &88.83  &91.00 & {\bf 97.19} &91.94 &{94.61} \\
 C12   &73.94 &96.06      &80.61 & 90.30&63.03 &46.06  &52.12 &64.55 &86.63 &{\bf 94.55} \\
 C13   &100.0 &100.0      &100.0 & 100.0&100.0 &97.78  &95.56 &97.78 &95.45 &{\bf 100.0} \\
 C14   &87.18 &89.74      &{97.44} & 92.31&66.67 &46.15  &48.72 &76.92 &82.05 &{\bf 100.0} \\
 C15   &100.0 &90.91      &100.0 & 100.0&100.0 &72.73  &81.82 &100.0 &100.0 &{\bf 100.0} \\
 C16   &100.0 &100.0      &100.0 & 100.0&100.0 &80.00  &100.0 &100.0 &100.0 &{\bf 100.0} \\
                                                                                                             
 \hline\hline
 {OA(\%)}   & 76.53   &  77.22   &82.97  & 78.72&81.11 &74.71 &65.16 &83.38  &80.29  &{\bf 89.40} \\
 
 {AA(\%)}   & 86.02   &  86.83   &90.19  & 88.15&84.97 &77.47 &70.21 &86.49  &86.61  &{\bf 94.11} \\
 
 {$\kappa$(\%)}&73.42 & 74.21    &80.65  & 75.81&78.51 &71.21 &60.26 &80.93  &77.40  &{\bf 87.85} \\
\bottomrule[1pt]
\end{tabular}
\end{center}
\vspace{-0.5cm}
\end{table*}

\begin{table*}[t]
\setlength{\aboverulesep}{-0.1pt}
\setlength{\belowrulesep}{-0.1pt}
\begin{center}
\caption{Classification accuracies (OA, AA, and $\kappa$) of different methods achieved on the Houston 2013 data.}
\label{table:houston2013_comparison}
%\vspace{-1ex}
\begin{tabular}{ c || c | c c c | c | c | c c c || c}
\toprule[1pt]
\multirow{2}{*}{Methods}     &  \multirow{2}{*}{SVM} &  \multicolumn{3}{c|}{CNNs} &  \multirow{2}{*}{RNN} & \multirow{2}{*}{miniGCN} & \multicolumn{3}{c||}{Transformers} &\multirow{2}{*}{HyperDID} \\
 \cmidrule(lr){3-5} \cmidrule(lr){8-10}
    &   &  { 3-D CNN}&  {PResNet }&  { HybridSN} &   &  & { ViT} & {SpectralFormer} & SSFTTNet & \\
\hline\hline
 C1   &  82.62  & 83.76  & 81.67&83.57 & 81.67&{\bf 96.20}  &82.53 &83.29 & 83.29& 82.15\\
 C2   &  98.78  & 95.49  & 99.91&{\bf 100.0} & 95.39&96.90  &99.06 &98.97 &90.51 & 99.53\\
 C3   &  90.30  & 95.05  & 90.89&98.02 & 95.05&{99.41}  &91.49 &96.63 &98.61 & {\bf 99.80}\\
 C4   &  97.06  & {99.24}  & 86.74&95.55 & 96.02&97.63  &95.64 &96.02 &96.97 & {\bf 99.91}\\
 C5   &  99.81  & 99.43  & 99.43&99.72 & 97.63&97.73  &99.34 &{\bf 100.0} &99.53 & 98.58\\
 C6   &  82.52  & 90.21  & 92.31&{95.80} & 91.61&95.10  &94.41 &94.40 &91.61 & {\bf 98.60}\\
 C7   &  89.65  & 86.85  & 90.49&90.67 & 89.92&65.86  &{\bf 91.04} &83.21 &67.91 & 90.11\\
 C8   &  57.74  & {82.05}  & 75.31&81.01 & 70.09&65.15  &60.68 &80.72 &55.08 & {\bf 83.95}\\
 C9   &  61.19  & 76.49  & {80.93}&{\bf 81.59} & 73.84&69.88  &71.20 &77.43 &54.25 & {70.54}\\
 C10   & 67.66   & 53.96  &70.27 &46.33 &65.93 &67.66  &52.51 &58.01 &{81.56} & {\bf 89.00}\\
 C11   & 72.68	   & 82.35  &84.91 &{\bf 94.12} &70.40 &82.83  &78.75 &80.27 &90.51 & 85.29\\
 C12   & 70.41   & 78.48  &71.85 &80.50 &79.73 & 68.40 &81.27 &84.44 &{84.73} & {\bf 84.73}\\
 C13   & 61.05   & 75.44  &89.47 &{\bf 94.74} &74.39 & 57.54 &65.96 &73.33 &81.75 & 92.98\\
 C14   & 94.33   & 91.90  &97.57 &96.36 &98.79 & 99.19 &95.14 &{\bf 99.60} &99.19 & 98.38\\
 C15   & 80.13   & 92.18  &{100.0} &95.78 &98.31 & 98.73 &92.39 &99.15 &99.79 & {\bf 100.0}\\
                                                                                                             
 \hline\hline
 {OA(\%)}      & 80.16  & 84.71 & 85.59 &86.89 &83.55 &82.31 &82.22 & 85.55 & 82.46 & {\bf 89.74}\\
 
 {AA(\%)}      & 80.40  & 85.53 & 87.45 &{88.92} &85.25 &83.88 &83.43 & 87.03 & 85.02 & {\bf 91.57}\\
  
 {$\kappa$(\%)} & 78.44   & 83.40 & 84.35 &85.77 &82.15 &80.84 &80.68 & 84.32 & 80.97 & {\bf 88.87}\\
\bottomrule[1pt]
\end{tabular}
\end{center}
\end{table*}

\section{Conclusions and Discussions}\label{sec:conclusions}

In this research endeavor, we introduce a groundbreaking hyperspectral intrinsic image decomposition method empowered by deep feature embedding, termed HyperDID. The comprehensive framework of HyperDID adeptly partitions hyperspectral images into environment-related and category-related features, thereby significantly enhancing classification performance. Comparative results underscore the efficacy of HyperDID in augmenting the representational capacity of deep models.
Our approach involves the development of key components such as the Environmental Feature Module (EFM) and Categorical Feature Module (CFM), both rooted in deep feature embedding principles to glean intrinsic features from hyperspectral images. Additionally, we introduce a Feature Discrimination Module (FDM) designed to segregate environment-related and category-related features. Ablation studies conducted validate the pivotal role of each module within HyperDID, affirming their collective effectiveness.
Detailed experiments provide compelling evidence for the efficacy and efficiency of HyperDID in the realm of hyperspectral image classification. By showcasing the method's ability to decompose and leverage intrinsic features, HyperDID emerges as a robust and innovative solution, poised to contribute significantly to advancing the field of hyperspectral image analysis.

Looking ahead, our research will delve into additional advanced strategies aimed at incorporating more physical characteristics and prior knowledge into the HyperDID framework to further enhance the decomposition of hyperspectral images. Beyond classification, we are keen to extend the application of HyperDID to other hyperspectral image processing tasks, such as anomaly detection and target identification.  Moreover, the exploration of alternative feature extraction frameworks based on the intrinsic characteristics of hyperspectral images is a promising avenue for future research.

% or


%\appendices

%\section{}

%% use section* for acknowledgment
%\section*{Acknowledgment}

%The authors would like to thank the researchers for providing the Pavia University, and the National Center for Airborne Laser Mapping and the Hyperspectral Image Analysis Laboratory at the University of Houston for acquiring and providing the data, and the IEEE GRSS Image Analysis and Data Fusion Technical Committee.

\ifCLASSOPTIONcaptionsoff
  \newpage
\fi



%\bibliographystyle{IEEEtran}

\bibliographystyle{IEEEtran}
\bibliography{egbib}

\begin{thebibliography}{10}
\providecommand{\url}[1]{#1}
\csname url@samestyle\endcsname
\providecommand{\newblock}{\relax}
\providecommand{\bibinfo}[2]{#2}
\providecommand{\BIBentrySTDinterwordspacing}{\spaceskip=0pt\relax}
\providecommand{\BIBentryALTinterwordstretchfactor}{4}
\providecommand{\BIBentryALTinterwordspacing}{\spaceskip=\fontdimen2\font plus
\BIBentryALTinterwordstretchfactor\fontdimen3\font minus
  \fontdimen4\font\relax}
\providecommand{\BIBforeignlanguage}[2]{{%
\expandafter\ifx\csname l@#1\endcsname\relax
\typeout{** WARNING: IEEEtran.bst: No hyphenation pattern has been}%
\typeout{** loaded for the language `#1'. Using the pattern for}%
\typeout{** the default language instead.}%
\else
\language=\csname l@#1\endcsname
\fi
#2}}
\providecommand{\BIBdecl}{\relax}
\BIBdecl

\bibitem{dong2022weighted}
Y.~Dong, Q.~Liu, B.~Du, and L.~Zhang, ``Weighted feature fusion of
  convolutional neural network and graph attention network for hyperspectral
  image classification,'' \emph{IEEE Transactions on Image Processing},
  vol.~31, pp. 1559--1572, 2022.

\bibitem{elmanawy2022hsi}
A.~I. ElManawy, D.~Sun, A.~Abdalla, Y.~Zhu, and H.~Cen, ``Hsi-pp: A flexible
  open-source software for hyperspectral imaging-based plant phenotyping,''
  \emph{Computers and Electronics in Agriculture}, vol. 200, p. 107248, 2022.

\bibitem{wang2022spectral}
Y.~Wang, D.~Hong, J.~Sha, L.~Gao, L.~Liu, Y.~Zhang, and X.~Rong,
  ``Spectral--spatial--temporal transformers for hyperspectral image change
  detection,'' \emph{IEEE Transactions on Geoscience and Remote Sensing},
  vol.~60, pp. 1--14, 2022.

\bibitem{jiao2023triplet}
J.~Jiao, Z.~Gong, and P.~Zhong, ``Triplet spectral-wise transformer network for
  hyperspectral target detection,'' \emph{IEEE Transactions on Geoscience and
  Remote Sensing}, 2023.

\bibitem{qu2020anomaly}
J.~Qu, Q.~Du, Y.~Li, L.~Tian, and H.~Xia, ``Anomaly detection in hyperspectral
  imagery based on gaussian mixture model,'' \emph{IEEE Transactions on
  Geoscience and Remote Sensing}, vol.~59, no.~11, pp. 9504--9517, 2020.

\bibitem{gong_manifold}
Z.~Gong, W.~Hu, X.~Du, P.~Zhong, and P.~Hu, ``Deep manifold embedding for
  hyperspectral image classification,'' \emph{IEEE Trans. Cybern.}, 2021.

\bibitem{b2}
Z.~Gong, P.~Zhong, and W.~Hu, ``Statistical loss and analysis for deep learning
  in hyperspectral image classification,'' \emph{IEEE Trans. Neural Netw.
  Learn. Syst.}, vol.~32, no.~1, pp. 322--333, 2021.

\bibitem{b1}
Z.~Gong, P.~Zhong, Y.~Yu, W.~Hu, and S.~Li, ``A cnn with multiscale convolution
  and diversified metric for hyperspectral image classification,'' \emph{IEEE
  Trans. Geosci. Remote Sens.}, vol.~57, no.~6, pp. 3599--3618, 2019.

\bibitem{petropoulos2012support}
G.~P. Petropoulos, C.~Kalaitzidis, and K.~P. Vadrevu, ``Support vector machines
  and object-based classification for obtaining land-use/cover cartography from
  hyperion hyperspectral imagery,'' \emph{Computers \& Geosciences}, vol.~41,
  pp. 99--107, 2012.

\bibitem{zhang2012application}
B.~Zhang, D.~Wu, L.~Zhang, Q.~Jiao, and Q.~Li, ``Application of hyperspectral
  remote sensing for environment monitoring in mining areas,''
  \emph{Environmental Earth Sciences}, vol.~65, pp. 649--658, 2012.

\bibitem{lu2020recent}
B.~Lu, P.~D. Dao, J.~Liu, Y.~He, and J.~Shang, ``Recent advances of
  hyperspectral imaging technology and applications in agriculture,''
  \emph{Remote Sensing}, vol.~12, no.~16, p. 2659, 2020.

\bibitem{roessner2001automated}
S.~Roessner, K.~Segl, U.~Heiden, and H.~Kaufmann, ``Automated differentiation
  of urban surfaces based on airborne hyperspectral imagery,'' \emph{IEEE
  Transactions on Geoscience and Remote sensing}, vol.~39, no.~7, pp.
  1525--1532, 2001.

\bibitem{kang2013feature}
X.~Kang, S.~Li, and J.~A. Benediktsson, ``Feature extraction of hyperspectral
  images with image fusion and recursive filtering,'' \emph{IEEE Transactions
  on Geoscience and Remote Sensing}, vol.~52, no.~6, pp. 3742--3752, 2013.

\bibitem{he2017recent}
L.~He, J.~Li, C.~Liu, and S.~Li, ``Recent advances on spectral--spatial
  hyperspectral image classification: An overview and new guidelines,''
  \emph{IEEE Transactions on Geoscience and Remote Sensing}, vol.~56, no.~3,
  pp. 1579--1597, 2017.

\bibitem{sun2014supervised}
L.~Sun, Z.~Wu, J.~Liu, L.~Xiao, and Z.~Wei, ``Supervised spectral--spatial
  hyperspectral image classification with weighted markov random fields,''
  \emph{IEEE Transactions on Geoscience and Remote Sensing}, vol.~53, no.~3,
  pp. 1490--1503, 2014.

\bibitem{guo2021improving}
A.~J. Guo and F.~Zhu, ``Improving deep hyperspectral image classification
  performance with spectral unmixing,'' \emph{Signal Processing}, vol. 183, p.
  107949, 2021.

\bibitem{lu2017subpixel}
T.~Lu, S.~Li, L.~Fang, X.~Jia, and J.~A. Benediktsson, ``From subpixel to
  superpixel: A novel fusion framework for hyperspectral image
  classification,'' \emph{IEEE Transactions on Geoscience and Remote Sensing},
  vol.~55, no.~8, pp. 4398--4411, 2017.

\bibitem{jin2017superpixel}
X.~Jin and Y.~Gu, ``Superpixel-based intrinsic image decomposition of
  hyperspectral images,'' \emph{IEEE Transactions on Geoscience and Remote
  Sensing}, vol.~55, no.~8, pp. 4285--4295, 2017.

\bibitem{xie2023hyperspectral}
W.~Xie, Y.~Gu, and T.~Liu, ``Hyperspectral intrinsic image decomposition based
  on physical prior driven unsupervised learning,'' \emph{IEEE Transactions on
  Geoscience and Remote Sensing}, 2023.

\bibitem{kang2015intrinsic}
X.~Kang, S.~Li, L.~Fang, and J.~A. Benediktsson, ``Intrinsic image
  decomposition for feature extraction of hyperspectral images,'' \emph{IEEE
  Transactions on Geoscience and Remote Sensing}, vol.~53, no.~4, pp.
  2241--2253, 2015.

\bibitem{jin2021intrinsic}
X.~Jin, Y.~Gu, and W.~Xie, ``Intrinsic hyperspectral image decomposition with
  dsm cues,'' \emph{IEEE Transactions on Geoscience and Remote Sensing},
  vol.~60, pp. 1--13, 2022.

\bibitem{gu2022hyperspectral}
Y.~Gu, W.~Xie, X.~Li, and X.~Jin, ``Hyperspectral intrinsic image decomposition
  with enhanced spatial information,'' \emph{IEEE Transactions on Geoscience
  and Remote Sensing}, vol.~60, pp. 1--14, 2022.

\bibitem{guo2008customizing}
B.~Guo, S.~R. Gunn, R.~I. Damper, and J.~D. Nelson, ``Customizing kernel
  functions for svm-based hyperspectral image classification,'' \emph{IEEE
  Transactions on Image Processing}, vol.~17, no.~4, pp. 622--629, 2008.

\bibitem{xia2017random}
J.~Xia, P.~Ghamisi, N.~Yokoya, and A.~Iwasaki, ``Random forest ensembles and
  extended multiextinction profiles for hyperspectral image classification,''
  \emph{IEEE Transactions on Geoscience and Remote Sensing}, vol.~56, no.~1,
  pp. 202--216, 2017.

\bibitem{kuching2007performance}
S.~Kuching, ``The performance of maximum likelihood, spectral angle mapper,
  neural network and decision tree classifiers in hyperspectral image
  analysis,'' \emph{Journal of Computer Science}, vol.~3, no.~6, pp. 419--423,
  2007.

\bibitem{ma2010local}
L.~Ma, M.~M. Crawford, and J.~Tian, ``Local manifold learning-based $ k
  $-nearest-neighbor for hyperspectral image classification,'' \emph{IEEE
  Transactions on Geoscience and Remote Sensing}, vol.~48, no.~11, pp.
  4099--4109, 2010.

\bibitem{xia2015random}
J.~Xia, M.~Dalla~Mura, J.~Chanussot, P.~Du, and X.~He, ``Random subspace
  ensembles for hyperspectral image classification with extended morphological
  attribute profiles,'' \emph{IEEE Transactions on Geoscience and Remote
  Sensing}, vol.~53, no.~9, pp. 4768--4786, 2015.

\bibitem{harsanyi1994hyperspectral}
J.~C. Harsanyi and C.-I. Chang, ``Hyperspectral image classification and
  dimensionality reduction: An orthogonal subspace projection approach,''
  \emph{IEEE Transactions on geoscience and remote sensing}, vol.~32, no.~4,
  pp. 779--785, 1994.

\bibitem{kuo2013kernel}
B.-C. Kuo, H.-H. Ho, C.-H. Li, C.-C. Hung, and J.-S. Taur, ``A kernel-based
  feature selection method for svm with rbf kernel for hyperspectral image
  classification,'' \emph{IEEE Journal of Selected Topics in Applied Earth
  Observations and Remote Sensing}, vol.~7, no.~1, pp. 317--326, 2013.

\bibitem{wang2023dynamic}
C.~Wang, L.~Zhang, W.~Wei, and Y.~Zhang, ``Dynamic super-pixel normalization
  for robust hyperspectral image classification,'' \emph{IEEE Transactions on
  Geoscience and Remote Sensing}, vol.~61, pp. 1--13, 2023.

\bibitem{lee2017going}
H.~Lee and H.~Kwon, ``Going deeper with contextual cnn for hyperspectral image
  classification,'' \emph{IEEE Transactions on Image Processing}, vol.~26,
  no.~10, pp. 4843--4855, 2017.

\bibitem{li2019deep}
S.~Li, W.~Song, L.~Fang, Y.~Chen, P.~Ghamisi, and J.~A. Benediktsson, ``Deep
  learning for hyperspectral image classification: An overview,'' \emph{IEEE
  Transactions on Geoscience and Remote Sensing}, vol.~57, no.~9, pp.
  6690--6709, 2019.

\bibitem{hang2019cascaded}
R.~Hang, Q.~Liu, D.~Hong, and P.~Ghamisi, ``Cascaded recurrent neural networks
  for hyperspectral image classification,'' \emph{IEEE Transactions on
  Geoscience and Remote Sensing}, vol.~57, no.~8, pp. 5384--5394, 2019.

\bibitem{mou2017deep}
L.~Mou, P.~Ghamisi, and X.~X. Zhu, ``Deep recurrent neural networks for
  hyperspectral image classification,'' \emph{IEEE Transactions on Geoscience
  and Remote Sensing}, vol.~55, no.~7, pp. 3639--3655, 2017.

\bibitem{ding2022multi}
Y.~Ding, Z.~Zhang, X.~Zhao, D.~Hong, W.~Cai, C.~Yu, N.~Yang, and W.~Cai,
  ``Multi-feature fusion: Graph neural network and cnn combining for
  hyperspectral image classification,'' \emph{Neurocomputing}, vol. 501, pp.
  246--257, 2022.

\bibitem{hong2020graph}
D.~Hong, L.~Gao, J.~Yao, B.~Zhang, A.~Plaza, and J.~Chanussot, ``Graph
  convolutional networks for hyperspectral image classification,'' \emph{IEEE
  Transactions on Geoscience and Remote Sensing}, vol.~59, no.~7, pp.
  5966--5978, 2020.

\bibitem{hong2021spectralformer}
D.~Hong, Z.~Han, J.~Yao, L.~Gao, B.~Zhang, A.~Plaza, and J.~Chanussot,
  ``Spectralformer: Rethinking hyperspectral image classification with
  transformers,'' \emph{IEEE Transactions on Geoscience and Remote Sensing},
  vol.~60, pp. 1--15, 2021.

\bibitem{zou2022lessformer}
J.~Zou, W.~He, and H.~Zhang, ``Lessformer: Local-enhanced spectral-spatial
  transformer for hyperspectral image classification,'' \emph{IEEE Transactions
  on Geoscience and Remote Sensing}, vol.~60, pp. 1--16, 2022.

\bibitem{roy2019hybridsn}
S.~K. Roy, G.~Krishna, S.~R. Dubey, and B.~B. Chaudhuri, ``Hybridsn: Exploring
  3-d--2-d cnn feature hierarchy for hyperspectral image classification,''
  \emph{IEEE Geoscience and Remote Sensing Letters}, vol.~17, no.~2, pp.
  277--281, 2019.

\bibitem{paoletti2018new}
M.~E. Paoletti, J.~M. Haut, J.~Plaza, and A.~Plaza, ``A new deep convolutional
  neural network for fast hyperspectral image classification,'' \emph{ISPRS
  journal of photogrammetry and remote sensing}, vol. 145, pp. 120--147, 2018.

\bibitem{gong2023deep}
Z.~Gong, X.~Zhou, and W.~Yao, ``Deep intrinsic decomposition with adversarial
  learning for hyperspectral image classification,'' \emph{arXiv preprint
  arXiv:2310.18549}, 2023.

\bibitem{wang2021contrastive}
P.~Wang, K.~Han, X.-S. Wei, L.~Zhang, and L.~Wang, ``Contrastive learning based
  hybrid networks for long-tailed image classification,'' in \emph{Proceedings
  of the IEEE/CVF conference on computer vision and pattern recognition}, 2021,
  pp. 943--952.

\bibitem{hou2021hyperspectral}
S.~Hou, H.~Shi, X.~Cao, X.~Zhang, and L.~Jiao, ``Hyperspectral imagery
  classification based on contrastive learning,'' \emph{IEEE Transactions on
  Geoscience and Remote Sensing}, vol.~60, pp. 1--13, 2021.

\bibitem{cao2021contrastnet}
Z.~Cao, X.~Li, Y.~Feng, S.~Chen, C.~Xia, and L.~Zhao, ``Contrastnet:
  Unsupervised feature learning by autoencoder and prototypical contrastive
  learning for hyperspectral imagery classification,'' \emph{Neurocomputing},
  vol. 460, pp. 71--83, 2021.

\bibitem{liu2023refined}
Q.~Liu, J.~Peng, Y.~Ning, N.~Chen, W.~Sun, Q.~Du, and Y.~Zhou, ``Refined
  prototypical contrastive learning for few-shot hyperspectral image
  classification,'' \emph{IEEE Transactions on Geoscience and Remote Sensing},
  vol.~61, pp. 1--14, 2023.

\bibitem{huang20223}
X.~Huang, M.~Dong, J.~Li, and X.~Guo, ``A 3-d-swin transformer-based
  hierarchical contrastive learning method for hyperspectral image
  classification,'' \emph{IEEE Transactions on Geoscience and Remote Sensing},
  vol.~60, pp. 1--15, 2022.

\bibitem{chen2020simple}
T.~Chen, S.~Kornblith, M.~Norouzi, and G.~Hinton, ``A simple framework for
  contrastive learning of visual representations,'' in \emph{International
  conference on machine learning}.\hskip 1em plus 0.5em minus 0.4em\relax PMLR,
  2020, pp. 1597--1607.

\bibitem{pavia}
``Hyperspectral data, accessed on nov. 16, 2023,'' https://www.ehu.ews/
  ccwintoco/index.php?title=Hyperspectral\_Remote\_Sensing\_Scenes.

\bibitem{houston2013}
``Houston 2013, accessed on nov. 16, 2023,''
  https://hyperspectral.ee.uh.edu/?page\_id=459.

\bibitem{paszke2019pytorch}
A.~Paszke, S.~Gross, F.~Massa, A.~Lerer, J.~Bradbury, G.~Chanan, T.~Killeen,
  Z.~Lin, N.~Gimelshein, L.~Antiga \emph{et~al.}, ``Pytorch: An imperative
  style, high-performance deep learning library,'' \emph{Advances in neural
  information processing systems}, vol.~32, 2019.

\bibitem{hamida20183}
A.~B. Hamida, A.~Benoit, P.~Lambert, and C.~B. Amar, ``3-d deep learning
  approach for remote sensing image classification,'' \emph{IEEE Transactions
  on geoscience and remote sensing}, vol.~56, no.~8, pp. 4420--4434, 2018.

\bibitem{paoletti2018deep}
M.~E. Paoletti, J.~M. Haut, R.~Fernandez-Beltran, J.~Plaza, A.~J. Plaza, and
  F.~Pla, ``Deep pyramidal residual networks for spectral--spatial
  hyperspectral image classification,'' \emph{IEEE Transactions on Geoscience
  and Remote Sensing}, vol.~57, no.~2, pp. 740--754, 2018.

\bibitem{dosovitskiy2020image}
A.~Dosovitskiy, L.~Beyer, A.~Kolesnikov, D.~Weissenborn, X.~Zhai,
  T.~Unterthiner, M.~Dehghani, M.~Minderer, G.~Heigold, S.~Gelly \emph{et~al.},
  ``An image is worth 16x16 words: Transformers for image recognition at
  scale,'' in \emph{International Conference on Learning Representations},
  2021.

\bibitem{sun2022spectral}
L.~Sun, G.~Zhao, Y.~Zheng, and Z.~Wu, ``Spectral--spatial feature tokenization
  transformer for hyperspectral image classification,'' \emph{IEEE Transactions
  on Geoscience and Remote Sensing}, vol.~60, pp. 1--14, 2022.

\end{thebibliography}

%\begin{thebibliography}{1}


%\end{thebibliography}



%\end{IEEEbiography}


\begin{IEEEbiographynophoto}{Zhiqiang Gong}
received the B.S. degree in applied mathematics from
Shanghai Jiao Tong University, Shanghai, China, in 2013, the M.S. degree
in applied mathematics from the National University of Defense Technology
(NUDT), Changsha, China, in 2015, and the Ph.D. degree in information and
communication engineering from the National Key Laboratory of Science and
Technology on ATR, NUDT, in 2019.

He is currently an Associate Professor with the Defense Innovation Institute, Chinese Academy of Military Sciences, Beijing, China.
He has authored more than 30 peer-reviewed articles in international journals, such as the IEEE TRANSACTIONS ON NEURAL NETWORKS AND LEARNING SYSTEMS, the IEEE TRANSACTIONS ON CYBERNETICS, the IEEE TRANSACTIONS ON GEOSCIENCE AND REMOTE SENSING, the SCIENCE CHINA INFORMATION SCIENCES, the IEEE GEOSCIENCE AND REMOTE SENSING LETTERS, and the IEEE JOURNAL OF SELECTED TOPICS IN APPLIED EARTH OBSERVATIONS AND REMOTE SENSING. His research interests are computer vision, machine learning, and image analysis.

He is a Referee of the IEEE TRANSACTIONS ON NEURAL NETWORKS AND LEARNING SYSTEMS, the IEEE TRANSACTIONS ON IMAGE PROCESSING, the IEEE TRANSACTIONS ON GEOSCIENCE AND REMOTE SENSING, the IEEE TRANSACTIONS ON INDUSTRIAL INFORMATICS, the IEEE JOURNAL OF SELECTED TOPICS IN APPLIED EARTH OBSERVATIONS AND REMOTE SENSING, and the IEEE GEOSCIENCE AND REMOTE SENSING LETTERS.
\end{IEEEbiographynophoto}
\begin{IEEEbiographynophoto}{Xian Zhou}
received the BE degree from the Department of Automation and Intelligent Science, Nankai University, Nankai, China in 2015 and the PhD degree from the Department of Computer Science and Engineering, Shanghai Jiao Tong University, Shanghai, China. 

She is currently an Associate Professor with the Information Research Center of Military Science, Chinese Academy of Military Sciences, Beijing, China. Her research interests include urban computing and data mining.
\end{IEEEbiographynophoto}

\begin{IEEEbiographynophoto}{Wen Yao}
received the M.Sc. and Ph.D. degrees in aerospace engineering from the National University of Defense Technology, Changsha, China, in 2007 and 2011, respectively. 

She is currently a Professor with the Defense Innovation Institute, Chinese Academy of Military Science, Beijing, China. Her research interests include spacecraft systems engineering, multidisciplinary design optimization, uncertainty-based optimization, and data-driven surrogate modeling and evolutionary optimization.
\end{IEEEbiographynophoto}

\begin{IEEEbiographynophoto}{Ping Zhong}
(Senior Member, IEEE) received the M.S. degree in applied mathematics and the Ph.D. degree in information and communication engineering from the National University of Defense Technology (NUDT), Changsha, China, in 2003 and 2008, respectively. From 2015 to 2016, he was a Visiting Scholar with the Department of Applied Mathematics and Theory Physics, University of Cambridge, Cambridge, U.K.

He is currently a Professor with the National Key Laboratory of Science and Technology on ATR, NUDT. He has authored more than 50 peer-reviewed papers in international journals such as the IEEE TRANSACTIONS ON NEURAL NETWORKS AND LEARNING SYSTEMS, the IEEE TRANSACTIONS ON IMAGE PROCESSING, the IEEE TRANSACTIONS ON GEOSCIENCE AND REMOTE SENSING, and the IEEE JOURNAL OF SELECTED TOPICS IN SIGNAL PROCESSING, and the IEEE JOURNAL OF SELECTED TOPICS IN APPLIED EARTH OBSERVATIONS AND REMOTE SENSING. His research interests include computer vision, machine learning, and pattern recognition.

Prof. Zhong was a recipient of the National Excellent Doctoral Dissertation Award of China in 2011 and the New Century Excellent Talents in University of China in 2013. He is a Referee of the IEEE TRANSACTIONS ON NEURAL METWORKS AND LEARNING SYSTEMS, the IEEE TRANSACTIONS ON IMAGE PROCESSING, IEEE TRANSACTIONS ON GEOSCIENCE AND REMOTE SENSING, the IEEE JOURNAL OF SELECTED TOPICS IN APPLIED EARTH OBSERVATIONS AND REMOTE SENSING, the IEEE JOURNAL OF SELECTED TOPICS IN SIGNAL PROCESSING, and the IEEE GEOSCIENCE AND REMOTE SENSING LETTERS.

\end{IEEEbiographynophoto}


\vfill


\end{document}